\pdfoutput=1

\documentclass[11pt]{article}

\usepackage[preprint]{acl}
\usepackage{makecell} 
\usepackage{float}
\usepackage{times}
\usepackage{latexsym}
\usepackage{booktabs}
\usepackage{caption}
\usepackage[T1]{fontenc}

\usepackage[utf8]{inputenc}

\usepackage{microtype}

\usepackage{inconsolata}

\usepackage{graphicx}

\usepackage{hyperref}

\usepackage{tcolorbox}
\tcbuselibrary{listingsutf8, breakable}

\usepackage{booktabs}
\usepackage{multirow}

\usepackage{graphicx}

\usepackage{soul} 

\title{ReDepress: A Cognitive Framework for Detecting Depression Relapse from Social Media}

\author{
 \textbf{Aakash Kumar Agarwal\textsuperscript{1}*},
 \textbf{Saprativa Bhattacharjee\textsuperscript{1}*},
 \textbf{Mauli Rastogi\textsuperscript{2}},
 \textbf{Jemima S. Jacob\textsuperscript{3}},
\\
 \textbf{Biplab Banerjee\textsuperscript{1}},
 \textbf{Rashmi Gupta\textsuperscript{1}} and
 \textbf{Pushpak Bhattacharyya\textsuperscript{1}}
\\
\\
 \textsuperscript{1}Indian Institute of Technology Bombay, Mumbai, India\\
 \textsuperscript{2}Clinical Psychologist, Bangalore, India \\
 \textsuperscript{3}Mental Health Consultant, HELPED, Mumbai, India
\\
\small{\texttt{\{aakash.agarwal, bbanerjee\}@iitb.ac.in, \{saprativa, pb\}@cse.iitb.ac.in,}}\\
\small{\texttt{\{maulirastogicp, jemimajohnjacob\}@gmail.com, rash\_cogsci@yahoo.com}}
}

\begin{document}
\maketitle
\def\thefootnote{*}\footnotetext{Equal contribution.}\def\thefootnote{\arabic{footnote}}

\begin{abstract}
Almost 50\% depression patients face the risk of going into relapse. The risk increases to 80\% after the second episode of depression. Although, depression detection from social media has attained considerable attention, depression relapse detection has remained largely unexplored due to the lack of curated datasets and the difficulty of distinguishing relapse and non-relapse users. In this work, we present \textit{ReDepress}, the first clinically validated social media dataset focused on relapse, comprising 204 Reddit users annotated by mental health professionals. Unlike prior approaches, our framework draws on cognitive theories of depression, incorporating constructs such as attention bias, interpretation bias, memory bias  and rumination into both annotation and modeling. Through statistical analyses and machine learning experiments, we demonstrate that cognitive markers significantly differentiate relapse and non-relapse groups, and that models enriched with these features achieve competitive performance, with transformer-based temporal models attaining an F1 of 0.86. Our findings validate psychological theories in real-world textual data and underscore the potential of cognitive-informed computational methods for early relapse detection, paving the way for scalable, low-cost interventions in mental healthcare.

\end{abstract}

\section{Introduction}
\label{sec:Introduction}

\begin{figure}
    \centering
    \includegraphics[width=1\linewidth]{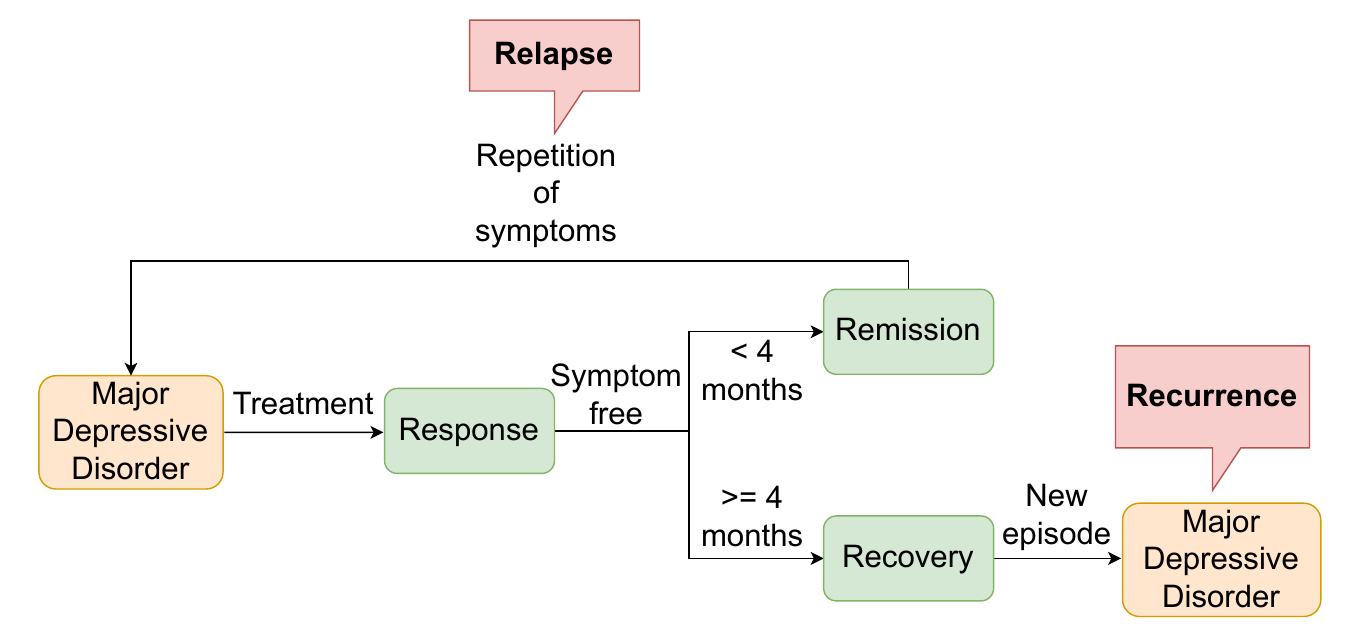}
    \caption{Visual depiction of relapse and other related terms encountered during the treatment of depression. The timelines are as per \citet{DBLP:conf/cbms/MuzammelOMS21}. In this work we use the terms relapse and recurrence synonymously. For a more elaborate discussion please see \autoref{app: Terminologies}.}
    \label{fig:relapse}
\end{figure}

In the treatment of depression, one of the biggest challenge faced by both medical practitioners and patients alike, is the risk of getting into relapse \cite{burcusa2007risk, American_Psychiatric_Association2025-yr}. Previous studies have found that, one out of every two depression patients (50\%), are likely to go into relapse. After the second episode of depression, this risk drastically increases to 80\%. Further, it becomes 90\% for those with a history of three episodes \cite{Kupfer1996-il, DBLP:conf/cbms/MuzammelOMS21}. This puts heavy burden on the patients as well as the healthcare infrastructure. Additionally, the risk of suicide among depression patients is twenty times higher than the normal populace \cite{Harris1997-hz}. Early detection of depression relapse thus becomes crucial for devising timely preventive intervention measures. 

Numerous computational studies have addressed the task of depression detection from social media posts either as a standalone disorder \cite{mendes-caseli-2024-identifying} or even in a setting with other comorbidities \cite{hengle-etal-2024-still}. Moreover, quite a few datasets are also available for depression detection from social media viz., e-Risk \cite{DBLP:conf/clef/LosadaCP18a} and Reddit Mental Health Dataset \cite{low2020natural}.


But there has been no study on the possibility of using social media posts for detection of depression relapse. This could be attributed to (1) the lack of a high quality depression relapse dataset, and (2) the inherent difficulty of separating relapse and non-relapse users because of the high similarity between the posts made by both set of users.

\begin{figure*}[!ht]
    \centering
    \includegraphics[width=1\linewidth]{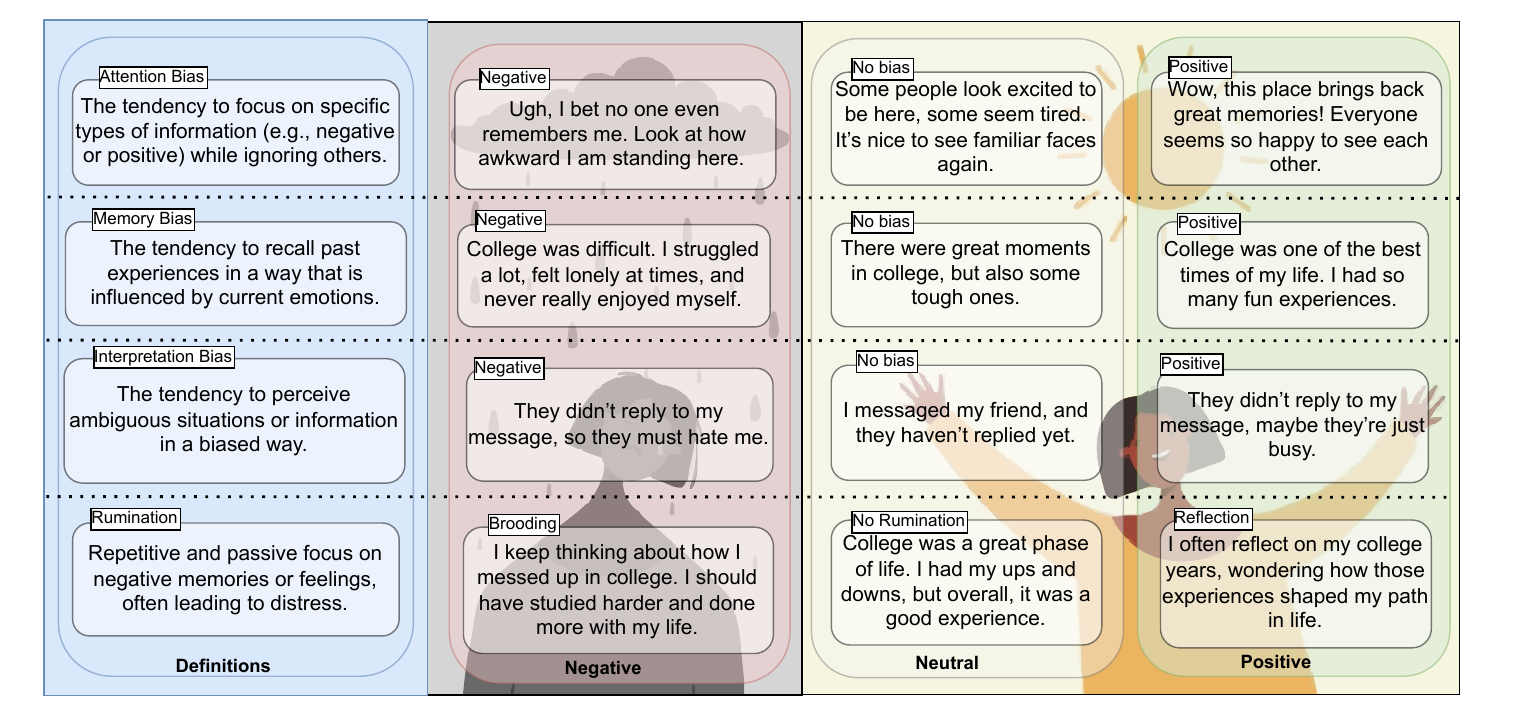}
    \caption{Definitions and examples of attention bias, memory bias, interpretation bias and rumination. Our ReDepress dataset is annotated on these cognitive dimensions (features) at the post level.}
    \label{fig:dimensions}
\end{figure*}

To address these gaps, in this work we introduce a high quality depression relapse dataset named \textit{ReDepress}, curated from the social media website Reddit. ReDepress is meticulously curated through a multistage process involving regular expressions, large language models (LLMs), non-expert humans and finally clinical psychologists. 

Studies in cognitive sciences have established the close connection between depression and cognitive processes such as rumination and information processing biases \cite{LeMoultGotlib2019, Gupta2012-bx, GotlibJoormann2010, Gupta2008-ho}. Moreover, the most widely used therapy for depression, Cognitive Behavior Therapy \cite{beck1967depression}, is also based on cognitive science. Thus, in this study we chose to study the effects of four cognitive constructs-- attention bias, memory bias, interpretation bias and rumination (see \autoref{fig:dimensions}). Our hypothesis is that, early identification of shifts in these biases and rumination could help detect relapse of depression. Moreover, we believe successful early detection of depression relapse based on social media posts could pave the way for a cost-effective approach to improve the quality of life for the depression patients and also ease the burden on the healthcare infrastructure.

Our contributions are:
\begin{enumerate}
    \item \textbf{ReDepress Dataset}: We present the first depression relapse focused social media dataset consisting of 204 (83 relapse and 121 non-relapse) users curated from Reddit. (\autoref{sec:Dataset})
    \item \textbf{Clinical Annotation}: Our dataset is annotated by three clinical psychologists who deal with depression patients regularly ensuring that our dataset is clinically validated from the very beginning. (\autoref{sec:Dataset})
    \item \textbf{Cognitive Markers for Relapse Detection}: We explore whether the four cognitive dimensions- attention bias, interpretation bias, memory bias and rumination correlate with relapse of depression, thereby validating psychological theories using real-world textual data. (\autoref{sec:Results and Discussion})
\end{enumerate}

\section{Related Work}
\label{sec:Related Work}

\subsection{Cognitive Theories of Depression}
Cognitive scientists underscore that depression is not only about low mood, but also about systematic biases in attention, interpretation and memory. \citet{Gotlib1984-wc} found that depressed individuals show a stable attentional bias toward depression-related words. \citet{butler1983cognitive} found that depressed individuals were more likely to endorse negative interpretations compared to non-depressed individuals. \citet{Lloyd1975-de} demonstrated that depressed individuals recall negative experiences faster than positive ones, linking memory bias to depression. Finally, \citet{Nolen-Hoeksema1991-cn} introduced rumination as a cognitive style that prolongs depressive episodes. These studies motivated our choice of picking the cognitive dimensions of attention bias, interpretation bias, memory bias and rumination for our work. 


\subsection{Clinical Studies on Depression Relapse}
\citet{burcusa2007risk} explores factors specifically associated with depression recurrence, differentiating them from those related to the initial onset of depression. \citet{DBLP:conf/medinfo/SalviniDLD15} proposes a multi-relational predictive model for depression relapse in bipolar disorder patients, employing Inductive Logic Programming (ILP) techniques on relational clinical data. \citet{Lye2020-eu} conducts a five-year longitudinal study of 201 patients with Major Depressive Disorder (MDD), examining a range of prognostic factors including clinical, personality, environmental, and genetic variables. Meanwhile, \citet{moriarty2021prognostic, moriarty2022predicting} provide a systematic review of existing prognostic models for depression relapse, highlighting significant methodological limitations and the widespread lack of external validation. Together, these studies highlight the high incidence of depression relapse and stress the urgent need for robust, clinically validated detection tools to facilitate personalized and timely interventions.


\subsection{Non-Clinical Studies on Depression Relapse}

\citet{DBLP:conf/ieaaie/AzizKT09} initiated research on modeling depression relapse through dynamic agent simulations, later extending this line of work into an integrative ambient agent framework that combined relapse dynamics with intelligent agent technology \citep{DBLP:journals/jaise/AzizKT10} for preventing relapse in individuals with a history of unipolar depression. Building on such conceptual models, subsequent efforts shifted toward data-driven prediction. For instance, \citet{DBLP:conf/kdd/NieGY16} leveraged the STAR*D dataset\footnote{\url{https://medicine.yale.edu/lab/statmethods/datasets/stard/}}
 and introduced a censored regression approach with truncated $l_{1}$ loss to estimate relapse risk across treatment stages. Similarly, \citet{Dwyer2019Modelling} explored the utility of machine learning, particularly LSTMs, on anonymized electronic health records (EHRs) for relapse forecasting.

Parallelly, researchers also examined signals from everyday digital traces. \citet{DBLP:conf/embc/GarciaHTFIY21} demonstrated that smartphone-derived lifelog data could be coupled with survival models to estimate relapse risk. Expanding further, \citet{DBLP:conf/ccscw/YinYWJLRD22} proposed an intelligent mobile monitoring platform that fused acoustic, semantic, environmental, and personal features via a CNN–LSTM model. More recently, wearable technologies have also been investigated, as in the work of \citet{Matcham2024-rc}, who studied the relationship between Fitbit-derived sleep features and relapse in individuals with recurrent depression. Complementing these, \citet{DBLP:conf/aaai/LucasiusAKBSS24} introduced a multimodal approach for adolescents by integrating video and speech data,  while \citet{DBLP:conf/cbms/MuzammelOMS21} incorporated audio-visual cues from the DAIC-WOZ dataset \citep{gratch-etal-2014-distress} into deep learning systems for detecting early signs of relapse.

Despite the breadth of methodologies, ranging from agent-based simulations to EHRs, smartphones, wearables and multimodal sensing, prior research has not explored the use of social media posts as a data source for relapse detection. We argue that analyzing social media posts offers a cost-effective and unobtrusive means of continuous monitoring, with the potential to enable early identification of vulnerable patients at scale.

\section{Dataset}
\label{sec:Dataset}

\begin{figure*}[!ht]
    \centering
    \includegraphics[width=0.9\linewidth]{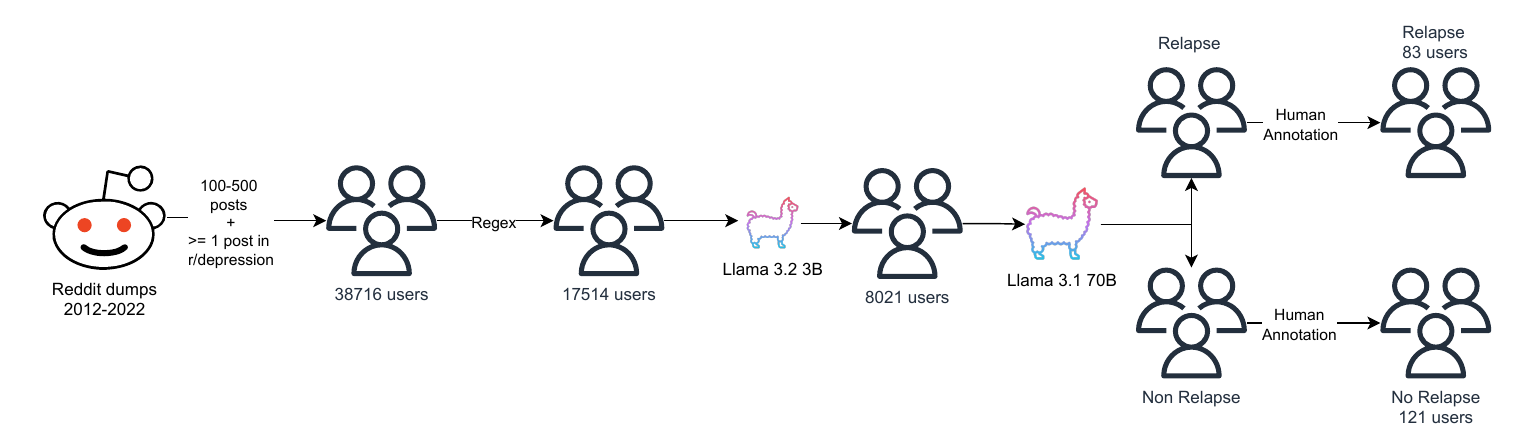}
    \caption{Dataset curation pipeline.}
    \label{fig:dataset}
\end{figure*}

\subsection{Data Curation}

Our dataset curation pipeline is depicted in \autoref{fig:dataset}. We start with a decade of Reddit dumps\footnote{Available at \url{https://the-eye.eu/redarcs/}} from 2012 to 2022 and consider only those users who have posted at least once in \textit{r/depression} subreddit and whose number of posts (overall) lie between 100 to 500. The lower limit is to have sufficient context per user and the upper limit is to minimize the cost and effort of manual annotation. For the first level of filtering we utilize an expanded version of the patterns provided by \citet{cohan-etal-2018-smhd} to extract the users who have self-reported being diagnosed with depression. We expanded the pattern list with the help of GPT-4o. The expanded list is provided in \autoref{app:patterns}. Next we use Llama 3.2 3B model on the posts of the users filtered in the previous step to further filter only those posts which has mention of any mental health issue to identify genuine depressed users and reduce false positives. We use a small model here because the number of input posts are quite high and it would incur a high computational cost if we use a bigger model. Also, we do not use GPT-4o here because of budget constraints. But smaller models are more prone to hallucinations, thus, finally we use Llama 3.1 70B model on the depression users identified by Llama 3.2 3B to further improve the accuracy. The prompts used are given in \autoref{app:prompts}.

\subsection{Timeline Extraction}
We create a dataset of Reddit user's timelines. A user's timeline is a subset of her entire posting history. Each user's posts are ordered chronologically. Depending on the answers of remission and relapse provided by Llama 3.1 70B, relapse and no relapse users are identified through manual inspection first by non-experts (first two authors). The manual inspection begins with verifying that the user has indeed self-reported being diagnosed with depression or not. Next, for creation of the timelines for a user, the post identified as remission is taken as the starting post. Then subsequent posts are checked up to a duration of one and a half year from the remission post. If a subsequent post (only those posts which had any answers given by Llama 3.1 70B as per prompt in \autoref{app:prompts}) mentioning either relapse or return of symptoms is found, the timeline is extracted and the user is a potential candidate belonging to the relapse class. Conversely, if no such posts are found for the specified duration, the user is most likely a no relapse case. Following this process we extract a total of 208 user timelines. We send these for annotation by clinical psychologists.

\subsection{Annotation}

\begin{table}[t]
\centering
\resizebox{1\columnwidth}{!}{%
\begin{tabular}{@{}lcccc@{}}
\toprule
 & \textbf{\#Users} & \textbf{Min \#Posts} & \textbf{Max \#Posts} & \textbf{Avg \#Posts}  \\ \midrule
Relapse     & 83 & 4  & 37  & 15.44  \\
No Relapse  & 121 & 3  & 35  & 11.24  \\ \bottomrule
\end{tabular}%
}
\caption{Dataset Statistics. Min \#Posts and Max \#Posts refer to the minimum and maximum number of posts of any user in that class respectively. Avg \#Posts refers to the average number of posts of all users in a class.}
\label{tab:dataset_stats}
\end{table}

\begin{table}
\centering
\resizebox{1\columnwidth}{!}{%
\begin{tabular}{@{}lccc@{}}
\toprule
 & \textbf{Full Agreement} & \textbf{Majority Agreement} & \textbf{Level} \\ \midrule
Attention Bias & 0.39 & 0.97 & Post level \\
Memory Bias & 0.18 & 0.98 & Post level \\
Interpretation Bias & 0.29 & 0.99 & Post level \\
Rumination & 0.17 & 0.92 & Post level \\ 
Relapse/No Relapse & 0.76 & 0.99 & User level \\ \bottomrule
\end{tabular}%
}
\caption{Inter-Annotator Agreement Metrics (Fleiss' Kappa). We follow \citet{tsakalidis-etal-2022-identifying} in reporting the majority agreement scores.}
\label{tab:agreement}
\end{table}

Final annotations are carried out by three clinical psychologists based on annotation guidelines (\autoref{app:Annotation Guidelines}) created after a series of discussion rounds. The annotations are carried out at two levels-- user level and post level. At the user level, the psychologists are asked to mark the remission and relapse posts of a user. Based on majority vote, i.e. if 2 out of 3 annotators mark any post of a given user as relapse, we consider that as a relapse user otherwise no relapse. Moreover, if the first couple of posts are not identified as a remission post, the user is discarded. Through this we get 83 relapse and 121 no relapse users. 4 users remained inconclusive due to lack of agreement with respect to either remission or relapse posts. The dataset statistics are presented in \autoref{tab:dataset_stats}.
\begin{figure}[t]
    \centering
    \includegraphics[width=1\linewidth]{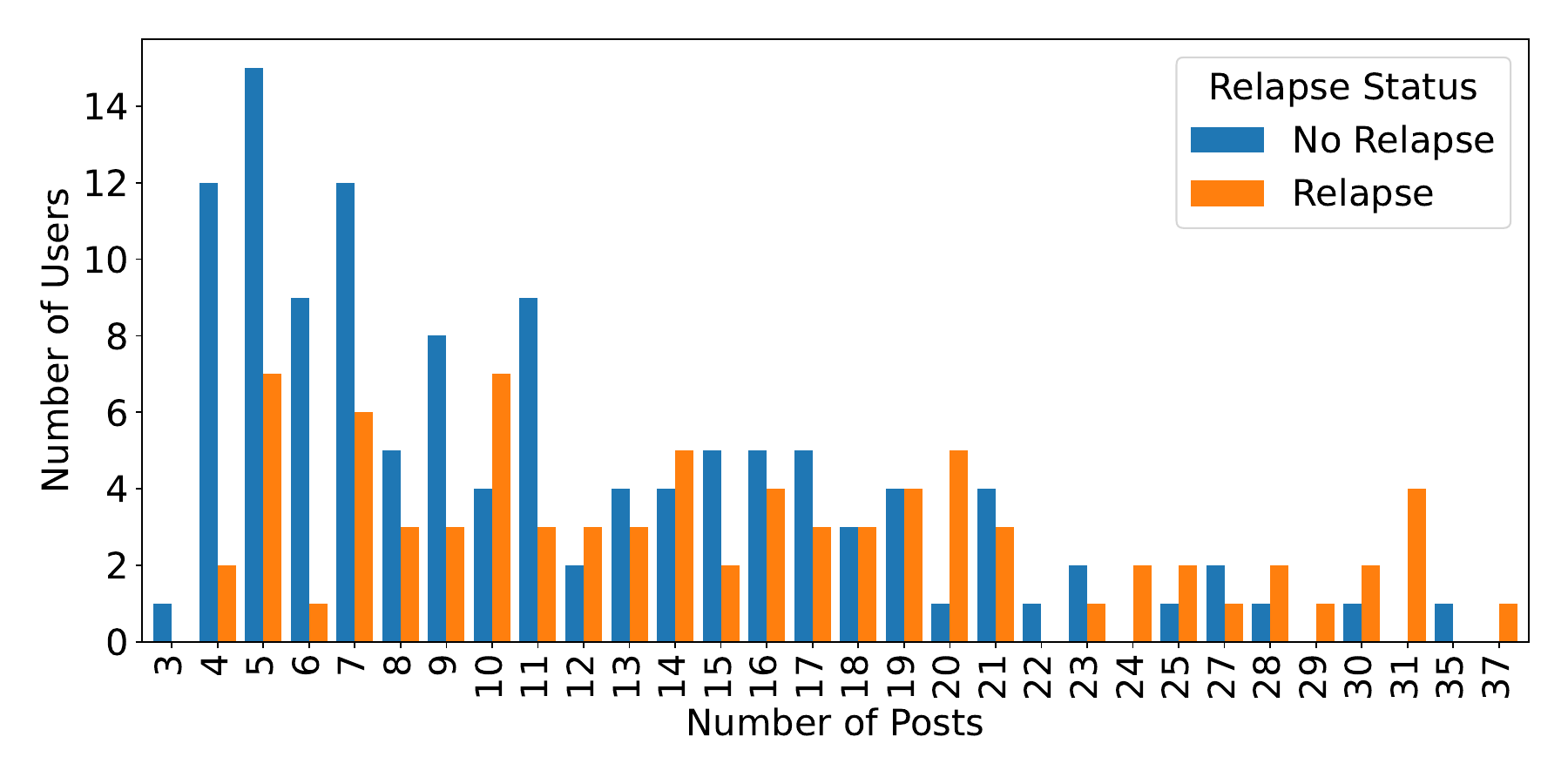}
    \caption{Distribution of user posts.}
    \label{fig:post_distribution}
\end{figure}
 
 At the post level, annotators are asked to annotate each post on four dimensions-- attention bias, interpretation bias, memory bias and rumination. The inter-annotator agreement metrics are presented in \autoref{tab:agreement}. The posts distributions are shown in \autoref{fig:post_distribution}. The three annotators employed by us were all female of the age group 20-30 years. All of them are clinical psychologists working at different organizations. The annotators were compensated for their efforts as per the norms. We plan to make the dataset available upon request under a user agreement to ensure responsible usage\footnote{Details available at \url{https://github.com/saprativa/ReDepress}}.

\subsection{Examples}

\begin{figure}[t]
    \centering
    \includegraphics[width=0.7\linewidth]{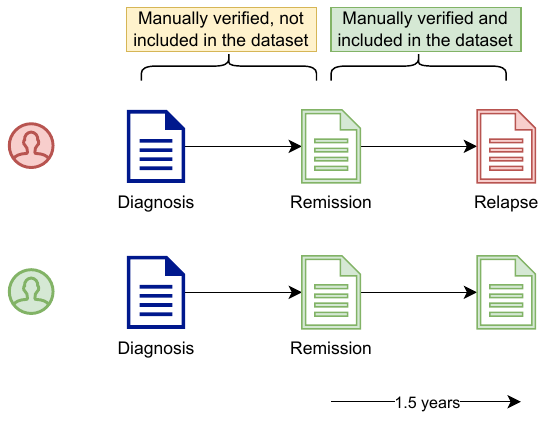}
    \caption{Timeline of a relapse vs no relapse user.}
    \label{fig:relapse_nonrelapse}
\end{figure}

An example timeline difference between relapse and no relapse is illustrated in \autoref{fig:relapse_nonrelapse}. Only the part from remission to relapse (or till one and a half years) is included in the dataset. The other part from diagnosis till remission (excluding) is only used during sample selection but not included in the final dataset. For a sample remission and relapse post from the dataset, please see \autoref{app:examples}. For an elaborate discussion on the how the distribution of relapse cases differs between our dataset based on Reddit and real life, please refer \autoref{app:distribution}.

\section{Experimental Setup}
\label{sec: Experimental Setup}

We conduct a series of experiments to evaluate the discriminative power of cognitive markers for relapse detection. At the user level, post-level annotations of attention bias, interpretation bias, memory bias, and rumination are aggregated using statistical measures such as mean, median, minimum, and maximum, and multiple machine learning classifiers are trained on these aggregated features. Hyperparameters are tuned via grid search, with evaluation performed using 5-fold cross-validation on 80\% of the data and the best-performing model tested on the remaining 20\%. 

To capture temporal dynamics in user timelines, we implement transformer-based encoders where each post is represented by a concatenation of its text embedding and cognitive markers, with the input being the chronologically ordered embeddings of each user. Finally, to test the feasibility of automating cognitive feature extraction, we train BERT-based classifiers for each cognitive dimension. We also benchmark large language models on this task in a zero-shot setting. Performance across all experiments are assessed with standard metrics including Accuracy, Precision, Recall, and F1-score.


\section{Results and Discussion}
\label{sec:Results and Discussion}

\subsection{Quantitative Analysis}

\begin{figure*}[!ht]
    \centering
    \includegraphics[width=1\linewidth]{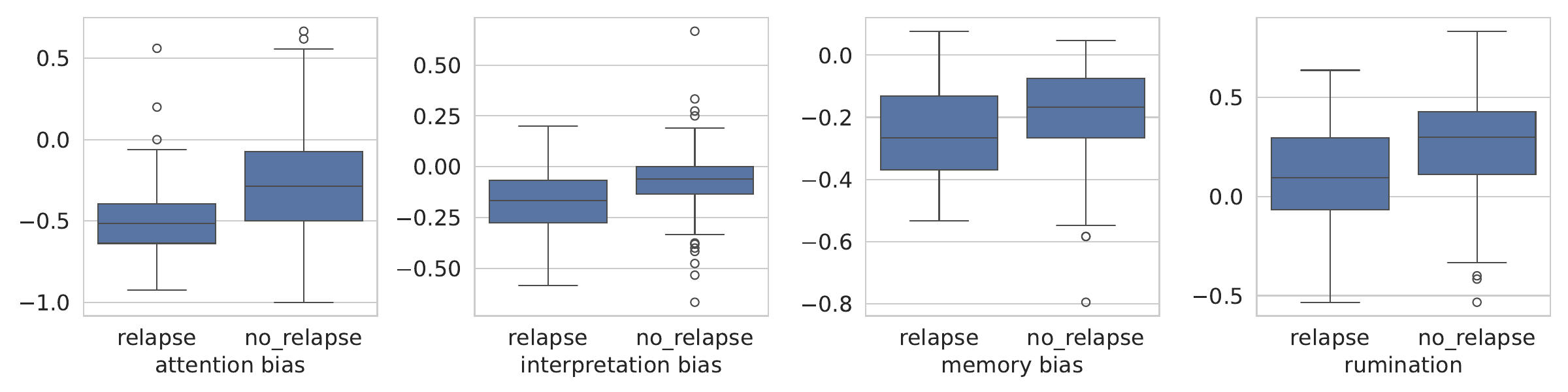}
    \caption{Distributions of average attention bias, interpretation bias, memory bias and rumination across relapse and no relapse groups. \autoref{app:average} discusses the averaging applied.}
    \label{fig:boxplot}
\end{figure*}

\begin{table}[t]
\centering
\resizebox{0.7\columnwidth}{!}{%
\begin{tabular}{@{}lc@{}}
\toprule
\textbf{Feature (Average)} & \textbf{Mann-Whitney p-value} \\ \midrule
Attention Bias & 0.000000 \\
Interpretation Bias & 0.000002 \\
Memory Bias & 0.000212 \\
Rumination & 0.000014 \\ \midrule
\end{tabular}%
}
\caption{Statistical test p-values comparing Relapse vs No Relapse groups for cognitive features (Average). To determine the appropriate statistical tests, we conducted normality tests on the variables. See \autoref{app:normality} for details.}
\label{tab:relapse_stats}
\end{table}

\autoref{fig:boxplot} displays the distribution of cognitive dimension scores-attention bias, interpretation bias, memory bias, and rumination-across relapse and no-relapse groups with relapse users exhibiting more negative attention, interpretation and memory biases, and also higher levels of brooding rumination. As evident from \autoref{tab:relapse_stats}, the differences between the two groups groups are statistically significant for all cognitive features ($p < 0.01$).  

\begin{figure}
\centering 
\includegraphics[width=1\linewidth]{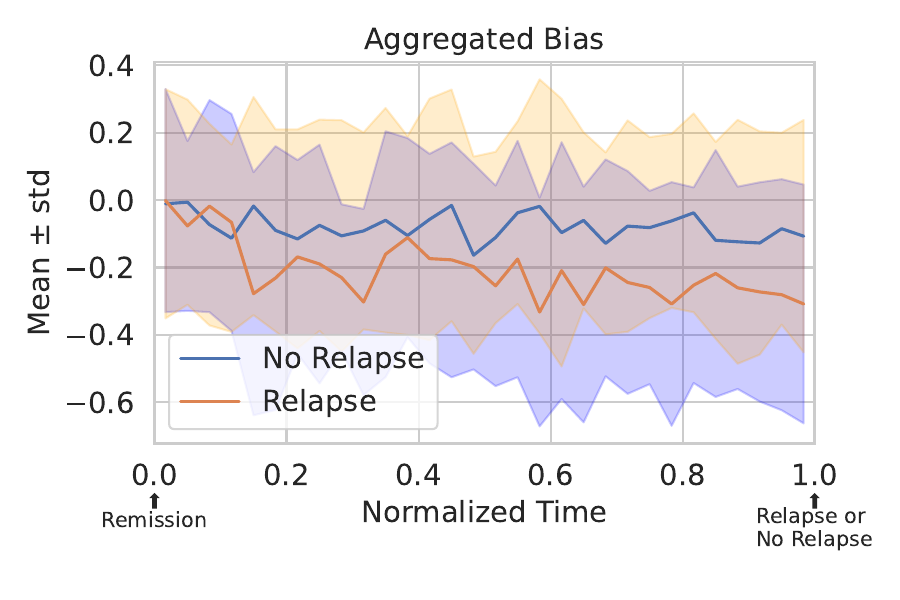} \caption{Temporal evolution of aggregated cognitive bias scores across normalized user timelines.} \label{fig:combined_all_biases_normalized} 
\end{figure}

In addition to static distributions, we also examined the temporal evolution of cognitive biases by aggregating multiple bias measures into combined scores and normalizing them over user timelines. \autoref{fig:combined_all_biases_normalized} illustrates how these combined bias scores change across time for relapse and no-relapse groups. Individuals who eventually relapsed exhibited a stronger downward trajectory, with mean values consistently below those of no-relapse group, indicating more persistent negative cognitive biases. While both groups displayed variability, the divergence between their trajectories became increasingly pronounced as time progressed, underscoring the temporal buildup of cognitive vulnerabilities in relapse cases.

\begin{figure}
    \centering
    \includegraphics[width=1\linewidth]{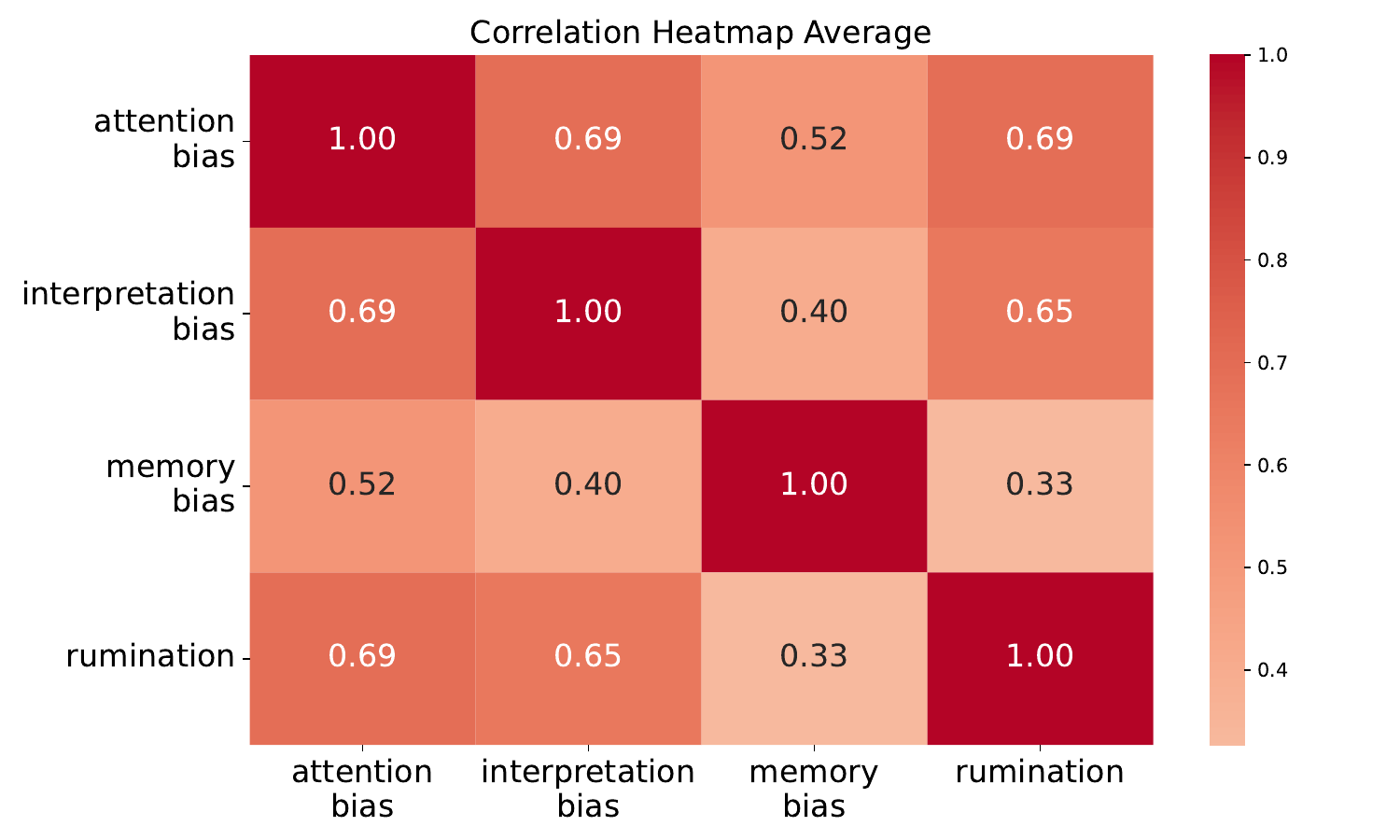}
    \caption{Cognitive dimensions correlation heatmap.}
    \label{fig:heatmap_dimensions}
\end{figure}

We also computed a correlation heatmap (\autoref{fig:heatmap_dimensions}) based on average user-level scores across the four annotated features. The analysis revealed moderately strong positive correlations between attention bias and interpretation bias (r = 0.69), as well as between attention bias and rumination (r = 0.69), and interpretation bias and rumination (r = 0.65). Memory bias showed weaker, though still positive, associations with the other dimensions, most notably with attention bias (r = 0.52). 

Importantly, none of the cognitive dimensions showed very high correlations (r > 0.80), suggesting that these measures capture related but distinct processes, rather than reflecting a single underlying factor such as sentiment alone. Overall, this correlation structure underscores the interconnected nature of cognitive vulnerabilities in depression and supports our hypothesis that a multidimensional cognitive lens is essential for understanding and detecting relapse.

These results are consistent with clinical studies, which demonstrate that individuals vulnerable to relapse tend to display persistent negative biases in attention, memory, and interpretation, as well as increased rumination, even after remission \cite{Visted2018-si, LeMoultGotlib2019}. However, as visible in the \autoref{fig:boxplot}, while group differences are significant, the distributions overlap substantially, reflecting the clinical reality that remitted individuals often retain residual cognitive vulnerabilities, though to a lesser degree than those in active relapse.

\subsection{Cognitive Feature-Only Modeling}
We computed cognitive bias features for each user by aggregating their post-level ratings using statistical measures (mean, median, min and max ) across the four cognitive dimensions: attention, interpretation, memory, and rumination. These aggregated features served as inputs for several machine learning models, including Random Forest, XGBoost, Logistic Regression, Gradient Boosting, SVM, KNN, and Neural Network.

\autoref{tab:updated_model_performance} reports the performance of each model. The GradientBoosting classifier achieved the best overall performance (Accuracy: 0.80, F1-Score: 0.78), indicating its effectiveness in distinguishing between relapse and non-relapse users based on the cognitive feature set. The best performing hyperparameters are listed in \autoref{tab:best_hyperparameters_with_aggregations_balancing} (\autoref{app:hyper}). Neural Network also performed competitively, with F1-score of 0.76. While these results are promising, the overall performance, with an F1-score of 0.78 at its best, is not exceptionally high. This suggests that the models could potentially benefit from additional features, such as those present within the text of the posts, to improve their discriminative power and better distinguish between relapse and non-relapse users.

\begin{table}[t]
\centering
\resizebox{1\columnwidth}{!}{%
\begin{tabular}{@{}lcccc@{}}
\toprule
\textbf{Model} & \textbf{Accuracy} & \textbf{Precision} & \textbf{Recall} & \textbf{F1} \\ \midrule
RandomForest        & 0.71 & 0.59 & \textbf{0.94} & 0.73 \\
XGBoost             & 0.73 & 0.62 & \textbf{0.94} & 0.74 \\
LogisticRegression  & 0.73 & 0.62 & \textbf{0.94} & 0.74 \\
GradientBoosting    & \textbf{0.80} & \textbf{0.74} & 0.82 & \textbf{0.78} \\
SVM                 & 0.73 & 0.62 & \textbf{0.94} & 0.74 \\
KNN                 & 0.68 & 0.57 & \textbf{0.94} & 0.71 \\
Neural Network       & 0.78 & 0.70 & 0.82 & 0.76 \\ \bottomrule
\end{tabular}%
}
\caption{Performance Metrics for Different Models on Test Set. Best values are in \textbf{bold}.}
\label{tab:updated_model_performance}
\end{table}

\subsection{ Interpretation and Clinical Relevance}
Our findings validate cognitive models of depression, showing that users at risk of relapse exhibit more pronounced negative attention, memory, and interpretation biases, as well as elevated rumination, mirroring cognitive patterns observed in clinical populations. The results suggest that cognitive features extracted from social media can meaningfully distinguish relapse from no-relapse cases, but also reinforce the subtlety of these differences: remitted individuals retain mild cognitive vulnerabilities, as reflected by the moderate but not extreme model performances and overlapping group distributions. This is in line with clinical literature indicating residual symptoms and biases often persist after remission, contributing to future relapse risk.

\subsection{Ablation study}
\label{subsec:feature_importance}

To further investigate the individual importance of each cognitive bias feature in relapse detection, we conducted comprehensive ablation studies using both model-free and model-based approaches.

\begin{table}[t]
\centering
\resizebox{1\columnwidth}{!}{%
\begin{tabular}{@{}l l c c c@{}}
\toprule
\textbf{Feature} & \textbf{Statistic} & \textbf{II$_\mathrm{Ablated}$} & \boldmath{$\Delta$} \textbf{II} & \textbf{Interpretation} \\
\midrule
Rumination          & Mean   & 0.611 & +0.014  & Most important \\
Interpretation Bias  & Mean   & 0.605 & +0.008  & Some importance                      \\
\bottomrule
\end{tabular}%
}
\caption{Top two features contributing most to relapse detection according to information imbalance ablation analysis. Higher positive $\Delta$II indicates greater unique informativeness.}
\label{tab:info_imbalance_top2}
\end{table}

\textbf{Model-Free Feature Importance: Information Imbalance Ablation}
We first employed a model-free, information-theoretic ablation based on the information imbalance metric\footnote{\url{https://dadapy.readthedocs.io/en/latest/jupyter_example_2.html}}. This method measures the change in information transfer from features to the relapse label when a given feature is removed, independent of any specific model. \autoref{tab:info_imbalance_top2} shows the top two most informative features, while \autoref{tab:info_imbalance_ablation} in \autoref{app:info_imbalance} provides the full ablation results for each type of bias and aggregation (min, max, mean, median). A larger positive $\Delta$II indicates a higher unique contribution of that feature to the label; negative or near-zero values suggest redundancy.

Results indicate that the mean of rumination ($+0.014$) and the mean of interpretation bias ($+0.008$) are the most critical features, as their removal leads to the greatest increase in information imbalance. In contrast, features such as the median of interpretation bias ($-0.025$) and the maximum of rumination and attention bias ($-0.020$) show negative or negligible $\Delta$II, suggesting these features are largely redundant or even introduce noise.

\begin{table}[t]
\centering
\resizebox{0.8\columnwidth}{!}{%
\begin{tabular}{@{}lcccc@{}}
\toprule
\textbf{Feature} & \textbf{Mean} & \textbf{Min} & \textbf{Max} & \textbf{Median} \\
\midrule
Memory Bias & 0.159 & 0.094 & 0.041 & 0.060 \\
Attention Bias & 0.078 & 0.041 & 0.111 & 0.060 \\
Rumination & 0.041 & 0.021 & 0.078 & 0.111 \\
Interpretation Bias & 0.009 & 0.060 & 0.095 & 0.078 \\
\bottomrule
\end{tabular}%
}
\caption{F1 drop per feature type and aggregation.}
\label{tab:f1_ablation}
\end{table}

\textbf{Model-Based Feature Importance: F1 Score Ablation}
To complement the model-free analysis, we performed a model-based ablation using the best-performing classifier (GradientBoosting classifier, F1-Score: 0.78). For each feature, we retrained the model after removing that feature and measured the drop in F1 score on the test set. The results are summarized in \autoref{tab:f1_ablation}.

The largest F1 drop occurs when removing Memory Bias, particularly the mean aggregation (0.159), followed by Attention Bias - Max (0.111) and Rumination - Median (0.111). Removing Interpretation Bias results in only a minimal change across aggregations (maximum F1 drop 0.095), suggesting that it contributes limited unique discriminative information. Overall, features related to memory and attention biases appear to have the strongest impact on the model’s discriminative performance.

\subsection{Temporal Sequence Modeling}

\begin{table}[ht]
\centering
\resizebox{1\linewidth}{!}{%
\begin{tabular}{@{}lccccc@{}}
\toprule
\textbf{Embedding} & \textbf{Strategy} & \textbf{Accuracy} & \textbf{Precision} & \textbf{Recall} & \textbf{F1} \\ \midrule
mentalbert & No CM & 0.81 & 0.85 & 0.81 & 0.80 \\
mentalbert & CM & \textbf{0.83} & \textbf{0.86} & \textbf{0.83} & \textbf{0.82} \\
mentalbert & CM-Emb & \textbf{0.83} & \textbf{0.86} & \textbf{0.83} & \textbf{0.82} \\ \midrule

mentalroberta & No CM & \textbf{0.83} & 0.85 & \textbf{0.83} & 0.80 \\
mentalroberta & CM & 0.81 & 0.83 & 0.81 & \textbf{0.81} \\
mentalroberta & CM-Emb & 0.81 & \textbf{0.87} & 0.81 & 0.79 \\ \midrule

mpnet & No CM & 0.86 & 0.86 & 0.86 & 0.81 \\
mpnet & CM & \textbf{0.88}$^{*}$ & 0.88 & \textbf{0.88}$^{*}$ & \textbf{0.86}$^{*}$ \\
mpnet & CM-Emb & \textbf{0.88}$^{*}$ & \textbf{0.89}$^{*}$ & \textbf{0.88}$^{*}$ & \textbf{0.86}$^{*}$ \\ 
\bottomrule
\end{tabular}%
}
\caption{Binary relapse detection results using transformer embeddings under three settings:  
No CM = No Cognitive Markers,  
CM = Cognitive Markers,  
CM-Emb = Cognitive Markers Embedding.  
\textbf{Bold} highlights the best result within each embedding type, and $^{*}$ marks the overall best across all embeddings.}
\label{tab:relapse_binary_extended}
\end{table}

While aggregated features provide useful signals, they ignore the sequential nature of user timelines. To model temporal dependencies, we implemented a transformer-based encoder where each Reddit post is represented by its text embedding concatenated with four cognitive markers (attention bias, interpretation bias, memory bias, rumination). The chronological sequences are encoded, and the final valid hidden state is used for classification. Unlike the original transformer \cite{10.5555/3295222.3295349} model that takes tokens as input, our model takes post embeddings as input . 

We experimented with three input configurations:
\begin{itemize}
    \item \textbf{No Cognitive Markers (No CM):} Embedding (\texttt{<post>})
    \item \textbf{Cognitive Markers (CM):} Embedding (\texttt{<post>}) + \texttt{<Cognitive Markers>}
    \item \textbf{Cognitive Markers Embedding (CM-Emb):} Embedding (\texttt{<post>} + \texttt{<Cognitive Markers>})
\end{itemize}

\autoref{tab:relapse_binary_extended} summarizes results across \texttt{mentalbert}, \texttt{mentalroberta} \cite{ji-etal-2022-mentalbert}, and \texttt{mpnet} \cite{DBLP:conf/nips/Song0QLL20}. Cognitive markers consistently enhance performance, with the largest gains observed for \texttt{mpnet}, where F1 improved from 0.81 (embeddings only) to 0.86 with markers. \texttt{mentalbert} also benefited, achieving 0.82 F1 with markers compared to 0.80 without.  

Interestingly, while mental health domain-adapted models such as \texttt{mentalbert} and \texttt{mentalroberta} perform well, they are consistently outperformed by \texttt{mpnet}, which is not domain-adapted. This suggests that domain-adapted models are primarily optimized for tasks with clear diagnostic differences, such as depression detection, whereas relapse detection on Redepress involves more subtle cognitive and behavioral cues.

Overall, these results show that temporal sequence models enriched with cognitive features better capture relapse risk, highlighting the importance of integrating clinical insights into sequential text modeling. The best hyperparameter settings for these transformer models are reported in \autoref{tab:best_hyperparameters_transformers} (\autoref{app:hyper}).

\subsection{Cognitive Dimension Classifiers}

For assessing the feasibility of automated cognitive dimension labeling, we trained dedicated classifiers for each cognitive feature using multiple transformer-based models (\autoref{tab:dim_classifier}). We performed three-class classification for each cognitive construct individually. Specifically, attention bias, memory bias, and interpretation bias were classified into \textit{positive}, \textit{negative}, and \textit{no bias} categories, while rumination was classified into \textit{reflection}, \textit{brooding}, and \textit{no rumination}. The full label distributions for these are provided in \autoref{tab:majority_distribution}.

\begin{table}[t]
\centering
\resizebox{\columnwidth}{!}{%
\begin{tabular}{@{}ccccc c@{}}
\toprule
\textbf{Feature} & \textbf{Model} & \textbf{Accuracy} & \textbf{Precision} & \textbf{Recall} & \textbf{F1} \\ \midrule
\multirow{4}{*}{Attention bias} 
 & bert-base-uncased & 0.66 & 0.68 & 0.66 & 0.66 \\
 & clinicalbert & 0.67 & 0.67 & 0.67 & 0.67 \\
 & mentalbert & \textbf{0.74} & \textbf{0.73} & \textbf{0.74} & \textbf{0.73} \\
 & mentalroberta & 0.73 & \textbf{0.73} & 0.73 & \textbf{0.73} \\ \midrule
\multirow{4}{*}{Interpretation bias} 
 & bert-base-uncased & 0.85 & 0.84 & 0.85 & 0.85 \\
 & clinicalbert & 0.85 & 0.83 & 0.85 & 0.84 \\
 & mentalbert & 0.87 & 0.87 & 0.87 & 0.87 \\
 & mentalroberta & \textbf{0.90} & \textbf{0.90} & \textbf{0.90} & \textbf{0.89} \\ \midrule
\multirow{4}{*}{Memory bias} 
 & bert-base-uncased & 0.80 & 0.78 & 0.80 & 0.79 \\
 & clinicalbert & 0.82 & 0.80 & 0.82 & 0.81 \\
 & mentalbert & 0.82 & 0.81 & 0.82 & 0.81 \\
 & mentalroberta & \textbf{0.83} & \textbf{0.82} & \textbf{0.83} & \textbf{0.82} \\ \midrule
\multirow{4}{*}{Rumination} 
 & bert-base-uncased & 0.72 & 0.72 & 0.72 & 0.72 \\
 & clinicalbert & 0.65 & 0.68 & 0.65 & 0.65 \\
 & mentalbert & 0.75 & 0.74 & 0.75 & 0.75 \\
 & mentalroberta & \textbf{0.76} & \textbf{0.76} & \textbf{0.76} & \textbf{0.76} \\ \bottomrule
\end{tabular}%
}
\caption{Results of cognitive dimension classifiers (best values in each block are highlighted in \textbf{bold}).}
\label{tab:dim_classifier}
\end{table}

\begin{table}[t]
\centering
\resizebox{0.8\columnwidth}{!}{%
\begin{tabular}{@{}lccc@{}}
\toprule
\textbf{Feature} & \textbf{Positive} & \textbf{Negative} & \textbf{No Bias} \\ \midrule
Memory Bias & 12 & 497 & 1828 \\
Attention Bias & 237 & 1201 & 899 \\
Interpretation Bias & 48 & 268 & 2021 \\ \midrule
\textbf{} & \textbf{Reflection} & \textbf{Brooding} & \textbf{No Rumination} \\
Rumination & 1216 & 576 & 545 \\ \bottomrule
\end{tabular}%
}
\caption{Final majority label distributions for each cognitive dimension.}
\label{tab:majority_distribution}
\end{table}


Results indicate that transformer models can reliably identify cognitive markers from text, with the best performance observed for interpretation and memory bias classification. Notably, the mental health–domain pretrained models (MentalBERT, MentalRoBERTa) outperform general-domain BERT variant \cite{DBLP:conf/naacl/DevlinCLT19} or even allied domain ClinicalBERT \cite{DBLP:journals/corr/abs-1904-05342}. This demonstrates that models adapted to mental health contexts are better at capturing nuanced cognitive constructs in social media posts. The relatively lower but consistent scores for attention bias and rumination suggest that these constructs are more challenging to capture, likely due to their subtler textual manifestations. Nevertheless, even for these dimensions, mental health–specific models achieve higher accuracy and F1 compared to their general-purpose counterparts.

\subsection{Zero-Shot Evaluation with LLMs}

In addition to fine-tuned classifiers, we also evaluated large language models (LLMs) in a zero-shot setting for each cognitive dimension (\autoref{tab:dim_classifier_llms}). The task setup was identical to the supervised experiments: attention bias, memory bias, and interpretation bias used the categories \textit{positive}, \textit{negative}, and \textit{no bias}, while rumination was classified into \textit{reflection}, \textit{brooding}, and \textit{no rumination}. The exact zero-shot prompt used for these evaluations is provided in \autoref{app:cognitive_prompt}.

Performance varied across models and dimensions. For attention bias, Llama-3.1-70B achieved the highest accuracy (0.72, F1 = 0.66), with Qwen-2-72B showing the highest precision (0.70). Interpretation bias was the most challenging (best: Llama-3.1-70B, Accuracy = 0.41, Recall = 0.63, F1 = 0.33). For memory bias, Qwen-2-72B performed best (Accuracy = 0.66, F1 = 0.47), while rumination was best captured by Llama-3.1-70B (Accuracy = 0.59, Recall = 0.60, F1 = 0.57). These results indicate that zero-shot LLMs can capture some cognitive dimensions without training, but their performance remains below domain-adapted supervised models, underscoring the need for careful annotation and domain-specific adaptation.

\begin{table}[t]
\centering
\resizebox{\columnwidth}{!}{%
\begin{tabular}{@{}ccccc c@{}}
\toprule
\textbf{Feature} & \textbf{Model} & \textbf{Accuracy} & \textbf{Precision} & \textbf{Recall} & \textbf{F1} \\ \midrule
\multirow{3}{*}{Attention bias} 
 & gemma3\_27b & 0.68 & 0.65 & 0.63 & 0.58 \\
 & llama3.1\_70b & \textbf{0.72} & 0.66 & \textbf{0.68} & \textbf{0.66} \\
 & qwen2\_72b & 0.71 & \textbf{0.70} & 0.63 & 0.65 \\ \midrule
\multirow{3}{*}{Interpretation bias} 
 & gemma3\_27b & 0.28 & \textbf{0.43} & 0.60 & 0.25 \\
 & llama3.1\_70b & \textbf{0.41} & \textbf{0.43} & \textbf{0.63} & \textbf{0.33} \\
 & qwen2\_72b & 0.35 & \textbf{0.43} & 0.58 & 0.30 \\ \midrule
\multirow{3}{*}{Memory bias} 
 & gemma3\_27b & 0.51 & 0.44 & \textbf{0.62} & 0.38 \\
 & llama3.1\_70b & 0.59 & 0.45 & \textbf{0.62} & 0.43 \\
 & qwen2\_72b & \textbf{0.66} & \textbf{0.48} & \textbf{0.62} & \textbf{0.47} \\ \midrule
\multirow{3}{*}{Rumination} 
 & gemma3\_27b & 0.56 & \textbf{0.58} & 0.58 & 0.54 \\
 & llama3.1\_70b & \textbf{0.59} & \textbf{0.58} & \textbf{0.60} & \textbf{0.57} \\
 & qwen2\_72b & 0.49 & 0.57 & 0.56 & 0.45 \\ \bottomrule
\end{tabular}%
}
\caption{Results of cognitive dimension classifiers for large language models (Gemma-3-27B \cite{gemmateam2025gemma3technicalreport}, Llama-3.1-70B \cite{grattafiori2024llama3herdmodels}, Qwen-2-72B \cite{yang2024qwen2technicalreport}). Best values in each block are highlighted in \textbf{bold}.}
\label{tab:dim_classifier_llms}
\end{table}

\section{Conclusion and Future Work}
\label{sec:Conclusion}
This paper introduced \textit{ReDepress}, the first clinically validated social media dataset dedicated to depression relapse, annotated by clinical psychologists and grounded in cognitive theory. Our analyses confirmed that cognitive markers (attention, interpretation, and memory biases alongside rumination) correlate strongly with relapse risk and can be computationally modeled from Reddit timelines. By integrating them into both annotation and modeling, we demonstrated that these markers are statistically significant indicators of relapse and capture subtle cognitive vulnerabilities reflected in user posts. Statistical tests, machine learning classifiers, and transformer-based sequence models further underscored their discriminative utility, with cognitive-enriched approaches consistently achieving performance gains. These results not only validate long-standing cognitive theories of depression but also highlight the potential of combining computational methods with clinical insights to support early relapse detection.

At the same time, our findings reveal the inherent challenges of the task: relapse and non-relapse users often share overlapping cognitive patterns, reflecting the subtle nature of residual vulnerabilities after remission. This underscores the need for more robust and context-aware approaches. Future work will focus on expanding ReDepress to larger and more diverse populations, incorporating multimodal signals such as behavioral and physiological data, and developing finer-grained temporal models that capture both short-term fluctuations and long-term cognitive shifts. We also envision personalized modeling frameworks that adapt to individual baselines and trajectories, increasing sensitivity to early relapse signs. Finally, ethical considerations around privacy, responsible deployment, and clinical translation remain central, requiring collaboration with mental health professionals to ensure safe and effective integration of such systems into real-world care.

\section*{Limitations}
\label{sec:Limitations}
While our work presents a novel approach to detecting depression relapse using social media data, it is essential to acknowledge its limitations.

\begin{itemize}
    \item \textbf{Annotation Subjectivity:}
    The identification of complex psychological constructs such as rumination or emotional distress relies on human annotation, which is inherently subjective. Despite our efforts to establish clear guidelines and conduct rigorous training, inter-annotator agreement remains a challenge. Variability in individual interpretations may introduce bias, particularly when labeling nuanced signals of relapse. 
    \item \textbf{External Validity:}
    Our dataset is derived exclusively from Reddit, which may limit the generalizability of our findings. Reddit users are not necessarily representative of the broader population, and their self-disclosures on the platform may differ from experiences shared in offline or clinical contexts. Furthermore, the self-reporting nature of social media introduces potential biases, as users may selectively share specific aspects of their mental health. Future research should explore datasets from diverse platforms or clinical sources to validate the robustness of our findings across contexts.
    \item \textbf{Temporal Granularity:}
    The temporal resolution of social media posts does not always align with clinical definitions of depression relapse. Users may post inconsistently, leading to gaps in data that obscure important transitions in mental state. Moreover, the timing of self-disclosures may lag behind or precede actual relapse episodes. To address this limitation, future studies could integrate multi-modal longitudinal data or use active user engagement methods to capture finer temporal details.
    \item \textbf{Ethical Considerations:}
    Analyzing sensitive mental health data from social media raises ethical concerns. While our data collection adheres to publicly available sources, there is an inherent risk of identifying or labeling vulnerable individuals. We prioritize user privacy by anonymizing data and following ethical research guidelines. Nevertheless, future work should consider additional safeguards, such as engaging with mental health professionals and developing frameworks to mitigate potential harm when applying these models in real-world scenarios.
\end{itemize}

\section*{Acknowledgments}
We would like to thank the reviewers for their valuable feedback, which helped shape this paper in the right direction. We also extend our gratitude to the members of the CFILT lab, IIT Bombay for their support and insightful discussions. Finally, we sincerely thank the annotators for their effort and contribution to this work.

\bibliography{custom}

\begin{thebibliography}{44}
\providecommand{\natexlab}[1]{#1}

\bibitem[{{American Psychiatric Association}(2025)}]{American_Psychiatric_Association2025-yr}
{American Psychiatric Association}. 2025.
\newblock \emph{Diagnostic and statistical manual of mental disorders, fifth edition, text revision ({DSM-5-TR\textregistered{}})}.
\newblock American Psychiatric Association Publishing, Arlington, TX.

\bibitem[{Aziz et~al.(2009)Aziz, Klein, and Treur}]{DBLP:conf/ieaaie/AzizKT09}
Azizi~Ab Aziz, Michel C.~A. Klein, and Jan Treur. 2009.
\newblock \href {https://doi.org/10.1007/978-3-642-02568-6\_4} {An agent model of temporal dynamics in relapse and recurrence in depression}.
\newblock In \emph{Next-Generation Applied Intelligence, 22nd International Conference on Industrial, Engineering and Other Applications of Applied Intelligent Systems, {IEA/AIE} 2009, Tainan, Taiwan, June 24-27, 2009. Proceedings}, volume 5579 of \emph{Lecture Notes in Computer Science}, pages 36--45. Springer.

\bibitem[{Aziz et~al.(2010)Aziz, Klein, and Treur}]{DBLP:journals/jaise/AzizKT10}
Azizi~Ab Aziz, Michel C.~A. Klein, and Jan Treur. 2010.
\newblock \href {https://doi.org/10.3233/AIS-2010-0054} {An integrative ambient agent model for unipolar depression relapse prevention}.
\newblock \emph{J. Ambient Intell. Smart Environ.}, 2(1):5--20.

\bibitem[{Beck(1967)}]{beck1967depression}
AT~Beck. 1967.
\newblock Depression: clinical, experimental, and theoretical aspects.
\newblock \emph{Harper \& Row google schola}, 2:103--113.

\bibitem[{Burcusa and Iacono(2007)}]{burcusa2007risk}
Stephanie~L Burcusa and William~G Iacono. 2007.
\newblock Risk for recurrence in depression.
\newblock \emph{Clinical psychology review}, 27(8):959--985.

\bibitem[{Butler and Mathews(1983)}]{butler1983cognitive}
Gillian Butler and Andrew Mathews. 1983.
\newblock Cognitive processes in anxiety.
\newblock \emph{Advances in behaviour research and therapy}, 5(1):51--62.

\bibitem[{Cohan et~al.(2018)Cohan, Desmet, Yates, Soldaini, MacAvaney, and Goharian}]{cohan-etal-2018-smhd}
Arman Cohan, Bart Desmet, Andrew Yates, Luca Soldaini, Sean MacAvaney, and Nazli Goharian. 2018.
\newblock \href {https://aclanthology.org/C18-1126/} {{SMHD}: a large-scale resource for exploring online language usage for multiple mental health conditions}.
\newblock In \emph{Proceedings of the 27th International Conference on Computational Linguistics}, pages 1485--1497, Santa Fe, New Mexico, USA. Association for Computational Linguistics.

\bibitem[{de~Zwart et~al.(2019)de~Zwart, Jeronimus, and de~Jonge}]{De_Zwart2019-pp}
P~L de~Zwart, B~F Jeronimus, and P~de~Jonge. 2019.
\newblock Empirical evidence for definitions of episode, remission, recovery, relapse and recurrence in depression: a systematic review.
\newblock \emph{Epidemiol. Psychiatr. Sci.}, 28(5):544--562.

\bibitem[{Devlin et~al.(2019)Devlin, Chang, Lee, and Toutanova}]{DBLP:conf/naacl/DevlinCLT19}
Jacob Devlin, Ming{-}Wei Chang, Kenton Lee, and Kristina Toutanova. 2019.
\newblock \href {https://doi.org/10.18653/V1/N19-1423} {{BERT:} pre-training of deep bidirectional transformers for language understanding}.
\newblock In \emph{Proceedings of the 2019 Conference of the North American Chapter of the Association for Computational Linguistics: Human Language Technologies, {NAACL-HLT} 2019, Minneapolis, MN, USA, June 2-7, 2019, Volume 1 (Long and Short Papers)}, pages 4171--4186. Association for Computational Linguistics.

\bibitem[{Dwyer(2019)}]{Dwyer2019Modelling}
Lesley Dwyer. 2019.
\newblock \href {https://github.com/LesleyDwyer/modelling-depression-recurrence/blob/master/Dwyer_DataScience_2019.pdf} {Modelling {Depression} {Recurrence} {Through} {Analysis} of {Electronic} {Health} {Records}}.
\newblock Technical report, City, University of London.

\bibitem[{Frank et~al.(1991)Frank, Prien, Jarrett, Keller, Kupfer, Lavori, Rush, and Weissman}]{frank_conceptualization_1991}
Ellen Frank, Robert~F. Prien, Robin~B. Jarrett, Martin~B. Keller, David~J. Kupfer, Philip~W. Lavori, A.~John Rush, and Myrna~M. Weissman. 1991.
\newblock \href {https://doi.org/10.1001/archpsyc.1991.01810330075011} {Conceptualization and rationale for consensus definitions of terms in major depressive disorder: {Remission}, recovery, relapse, and recurrence}.
\newblock \emph{Archives of General Psychiatry}, 48(9):851--855.

\bibitem[{Garcia et~al.(2021)Garcia, Hirao, Tajika, Furukawa, Ikeda, and Yoshimoto}]{DBLP:conf/embc/GarciaHTFIY21}
Felan Carlo~C. Garcia, Ayumi Hirao, Aran Tajika, Toshi~A. Furukawa, Kazushi Ikeda, and Junichiro Yoshimoto. 2021.
\newblock \href {https://doi.org/10.1109/EMBC46164.2021.9629798} {Leveraging longitudinal lifelog data using survival models for predicting risk of relapse among patients with depression in remission}.
\newblock In \emph{43rd Annual International Conference of the {IEEE} Engineering in Medicine {\&} Biology Society, {EMBC} 2021, Mexico, November 1-5, 2021}, pages 2455--2458. {IEEE}.

\bibitem[{Gotlib and Joormann(2010)}]{GotlibJoormann2010}
Ian~H. Gotlib and Jutta Joormann. 2010.
\newblock Cognition and depression: Current status and future directions.
\newblock \emph{Annual Review of Clinical Psychology}, 6:285--312.

\bibitem[{Gotlib and McCann(1984)}]{Gotlib1984-wc}
Ian~H Gotlib and C~Douglas McCann. 1984.
\newblock Construct accessibility and depression: An examination of cognitive and affective factors.
\newblock \emph{J. Pers. Soc. Psychol.}, 47(2):427--439.

\bibitem[{Gratch et~al.(2014)Gratch, Artstein, Lucas, Stratou, Scherer, Nazarian, Wood, Boberg, DeVault, Marsella, Traum, Rizzo, and Morency}]{gratch-etal-2014-distress}
Jonathan Gratch, Ron Artstein, Gale Lucas, Giota Stratou, Stefan Scherer, Angela Nazarian, Rachel Wood, Jill Boberg, David DeVault, Stacy Marsella, David Traum, Skip Rizzo, and Louis-Philippe Morency. 2014.
\newblock \href {https://aclanthology.org/L14-1421/} {The distress analysis interview corpus of human and computer interviews}.
\newblock In \emph{Proceedings of the Ninth International Conference on Language Resources and Evaluation ({LREC}'14)}, pages 3123--3128, Reykjavik, Iceland. European Language Resources Association (ELRA).

\bibitem[{Grattafiori et~al.(2024)Grattafiori, Dubey, Jauhri, Pandey, Kadian, and Al-Dahle}]{grattafiori2024llama3herdmodels}
Aaron Grattafiori, Abhimanyu Dubey, Abhinav Jauhri, Abhinav Pandey, Abhishek Kadian, and Ahmad Al-Dahle. 2024.
\newblock \href {https://arxiv.org/abs/2407.21783} {The llama 3 herd of models}.
\newblock \emph{Preprint}, arXiv:2407.21783.

\bibitem[{Gupta and Kar(2008)}]{Gupta2008-ho}
Rashmi Gupta and Bhoomika~R Kar. 2008.
\newblock Interpretive bias: Indicators of cognitive vulnerability to depression.
\newblock \emph{Ger. J. Psychiatr.}, 11(3):98--102.

\bibitem[{Gupta and Kar(2012)}]{Gupta2012-bx}
Rashmi Gupta and Bhoomika~R Kar. 2012.
\newblock Attention and memory biases as stable abnormalities among currently depressed and currently remitted individuals with unipolar depression.
\newblock \emph{Front. Psychiatry}, 3:99.

\bibitem[{Harris and Barraclough(1997)}]{Harris1997-hz}
E~Clare Harris and Brian Barraclough. 1997.
\newblock Suicide as an outcome for mental disorders.
\newblock \emph{Br. J. Psychiatry}, 170(3):205--228.

\bibitem[{Hengle et~al.(2024)Hengle, Kulkarni, Patankar, Chandrasekaran, D{'}silva, Jacob, and Gupta}]{hengle-etal-2024-still}
Amey Hengle, Atharva Kulkarni, Shantanu~Deepak Patankar, Madhumitha Chandrasekaran, Sneha D{'}silva, Jemima~S. Jacob, and Rashmi Gupta. 2024.
\newblock \href {https://doi.org/10.18653/v1/2024.emnlp-main.931} {Still not quite there! evaluating large language models for comorbid mental health diagnosis}.
\newblock In \emph{Proceedings of the 2024 Conference on Empirical Methods in Natural Language Processing}, pages 16698--16721, Miami, Florida, USA. Association for Computational Linguistics.

\bibitem[{Huang et~al.(2019)Huang, Altosaar, and Ranganath}]{DBLP:journals/corr/abs-1904-05342}
Kexin Huang, Jaan Altosaar, and Rajesh Ranganath. 2019.
\newblock \href {https://arxiv.org/abs/1904.05342} {Clinicalbert: Modeling clinical notes and predicting hospital readmission}.
\newblock \emph{CoRR}, abs/1904.05342.

\bibitem[{Ji et~al.(2022)Ji, Zhang, Ansari, Fu, Tiwari, and Cambria}]{ji-etal-2022-mentalbert}
Shaoxiong Ji, Tianlin Zhang, Luna Ansari, Jie Fu, Prayag Tiwari, and Erik Cambria. 2022.
\newblock \href {https://aclanthology.org/2022.lrec-1.778/} {{M}ental{BERT}: Publicly available pretrained language models for mental healthcare}.
\newblock In \emph{Proceedings of the Thirteenth Language Resources and Evaluation Conference}, pages 7184--7190, Marseille, France. European Language Resources Association.

\bibitem[{Kupfer et~al.(1996)Kupfer, Frank, and Wamhoff}]{Kupfer1996-il}
D~J Kupfer, E~Frank, and J~Wamhoff. 1996.
\newblock Mood disorders: Update on prevention of recurrence.
\newblock In C~Mundt, M~J Goldstein, K~Hahlweg, and P~Fiedler, editors, \emph{Interpersonal factors in the origin and course of affective disorders}, pages 289--302. Gaskell/Royal College of Psychiatrists.

\bibitem[{LeMoult and Gotlib(2019)}]{LeMoultGotlib2019}
Joelle LeMoult and Ian~H. Gotlib. 2019.
\newblock Depression: A cognitive perspective.
\newblock \emph{Clinical Psychology Review}, 69:51--66.

\bibitem[{Lloyd and Lishman(1975)}]{Lloyd1975-de}
G~G Lloyd and W~A Lishman. 1975.
\newblock Effect of depression on the speed of recall of pleasant and unpleasant experiences.
\newblock \emph{Psychol. Med.}, 5(2):173--180.

\bibitem[{Losada et~al.(2018)Losada, Crestani, and Parapar}]{DBLP:conf/clef/LosadaCP18a}
David~E. Losada, Fabio Crestani, and Javier Parapar. 2018.
\newblock \href {https://doi.org/10.1007/978-3-319-98932-7\_30} {Overview of erisk: Early risk prediction on the internet}.
\newblock In \emph{Experimental {IR} Meets Multilinguality, Multimodality, and Interaction - 9th International Conference of the {CLEF} Association, {CLEF} 2018, Avignon, France, September 10-14, 2018, Proceedings}, volume 11018 of \emph{Lecture Notes in Computer Science}, pages 343--361. Springer.

\bibitem[{Low et~al.(2020)Low, Rumker, Torous, Cecchi, Ghosh, and Talkar}]{low2020natural}
Daniel~M Low, Laurie Rumker, John Torous, Guillermo Cecchi, Satrajit~S Ghosh, and Tanya Talkar. 2020.
\newblock Natural language processing reveals vulnerable mental health support groups and heightened health anxiety on reddit during covid-19: Observational study.
\newblock \emph{Journal of medical Internet research}, 22(10):e22635.

\bibitem[{Lucasius et~al.(2024)Lucasius, Ali, Kundur, Battaglia, Szatmari, and Strauss}]{DBLP:conf/aaai/LucasiusAKBSS24}
Christopher Lucasius, Mai Ali, Deepa Kundur, Marco Battaglia, Peter Szatmari, and John Strauss. 2024.
\newblock \href {https://ceur-ws.org/Vol-3649/Paper22.pdf} {Prediction of relapse in adolescent depression using fusion of video and speech data (short paper)}.
\newblock In \emph{Proceedings of Machine Learning for Cognitive and Mental Health Workshop {(ML4CMH} 2024) Co-located with the Thirty-Eighth {AAAI} Conference on Artificial Intelligence {(AAAI} 2024), Vancouver, BC, Canada, February 26th, 2024}, volume 3649 of \emph{{CEUR} Workshop Proceedings}, pages 112--117. CEUR-WS.org.

\bibitem[{Lye et~al.(2020)Lye, Tey, Tor, Shahabudin, Ibrahim, Ling, Stanslas, Loh, Rosli, Lokman, Badamasi, Faris-Aldoghachi, and Abdul~Razak}]{Lye2020-eu}
Munn-Sann Lye, Yin-Yee Tey, Yin-Sim Tor, Aisya~Farhana Shahabudin, Normala Ibrahim, King-Hwa Ling, Johnson Stanslas, Su-Peng Loh, Rozita Rosli, Khairul~Aiman Lokman, Ibrahim~Mohammed Badamasi, Asraa Faris-Aldoghachi, and Nurul~Asyikin Abdul~Razak. 2020.
\newblock Predictors of recurrence of major depressive disorder.
\newblock \emph{PLoS One}, 15(3):e0230363.

\bibitem[{Matcham et~al.(2024)Matcham, Carr, Meyer, White, Oetzmann, Leightley, Lamers, Siddi, Cummins, Annas, de~Girolamo, Haro, Lavelle, Li, Lombardini, Mohr, Narayan, Penninx, Coromina, Riquelme~Alacid, Simblett, Nica, Wykes, Brasen, Myin-Germeys, Dobson, Folarin, Ranjan, Rashid, Dineley, Vairavan, Hotopf, and {RADAR-CNS consortium}}]{Matcham2024-rc}
F~Matcham, E~Carr, N~Meyer, K~M White, C~Oetzmann, D~Leightley, F~Lamers, S~Siddi, N~Cummins, P~Annas, G~de~Girolamo, J~M Haro, G~Lavelle, Q~Li, F~Lombardini, D~C Mohr, V~A Narayan, B~W H~J Penninx, M~Coromina, G~Riquelme~Alacid, S~K Simblett, R~Nica, T~Wykes, J~C Brasen, I~Myin-Germeys, R~J~B Dobson, A~A Folarin, Y~Ranjan, Z~Rashid, J~Dineley, S~Vairavan, M~Hotopf, and {RADAR-CNS consortium}. 2024.
\newblock The relationship between wearable-derived sleep features and relapse in major depressive disorder.
\newblock \emph{J. Affect. Disord.}, 363:90--98.

\bibitem[{Mendes and Caseli(2024)}]{mendes-caseli-2024-identifying}
Augusto~R. Mendes and Helena Caseli. 2024.
\newblock \href {https://aclanthology.org/2024.lrec-main.754/} {Identifying fine-grained depression signs in social media posts}.
\newblock In \emph{Proceedings of the 2024 Joint International Conference on Computational Linguistics, Language Resources and Evaluation (LREC-COLING 2024)}, pages 8594--8604, Torino, Italia. ELRA and ICCL.

\bibitem[{Moriarty et~al.(2021)Moriarty, Meader, Snell, Riley, Paton, Chew-Graham, Gilbody, Churchill, Phillips, Ali, and McMillan}]{moriarty2021prognostic}
Andrew~S Moriarty, Nicholas Meader, Kym~Ie Snell, Richard~D Riley, Lewis~W Paton, Carolyn~A Chew-Graham, Simon Gilbody, Rachel Churchill, Robert~S Phillips, Shehzad Ali, and Dean McMillan. 2021.
\newblock Prognostic models for predicting relapse or recurrence of major depressive disorder in adults.
\newblock \emph{Cochrane Database Syst. Rev.}, 5(5):CD013491.

\bibitem[{Moriarty et~al.(2022)Moriarty, Meader, Snell, Riley, Paton, Dawson, Hendon, Chew-Graham, Gilbody, Churchill et~al.}]{moriarty2022predicting}
Andrew~S Moriarty, Nicholas Meader, Kym~IE Snell, Richard~D Riley, Lewis~W Paton, Sarah Dawson, Jessica Hendon, Carolyn~A Chew-Graham, Simon Gilbody, Rachel Churchill, et~al. 2022.
\newblock Predicting relapse or recurrence of depression: systematic review of prognostic models.
\newblock \emph{The British Journal of Psychiatry}, 221(2):448--458.

\bibitem[{Muzammel et~al.(2021)Muzammel, Othmani, Mukherjee, and Salam}]{DBLP:conf/cbms/MuzammelOMS21}
Muhammad Muzammel, Alice Othmani, Himadri Mukherjee, and Hanan Salam. 2021.
\newblock \href {https://doi.org/10.1109/CBMS52027.2021.00018} {Identification of signs of depression relapse using audio-visual cues: {A} preliminary study}.
\newblock In \emph{34th {IEEE} International Symposium on Computer-Based Medical Systems, {CBMS} 2021, Aveiro, Portugal, June 7-9, 2021}, pages 62--67. {IEEE}.

\bibitem[{Nie et~al.(2016)Nie, Gong, and Ye}]{DBLP:conf/kdd/NieGY16}
Zhi Nie, Pinghua Gong, and Jieping Ye. 2016.
\newblock \href {https://doi.org/10.1145/2939672.2939870} {Predict risk of relapse for patients with multiple stages of treatment of depression}.
\newblock In \emph{Proceedings of the 22nd {ACM} {SIGKDD} International Conference on Knowledge Discovery and Data Mining, San Francisco, CA, USA, August 13-17, 2016}, pages 1795--1804. {ACM}.

\bibitem[{Nolen-Hoeksema(1991)}]{Nolen-Hoeksema1991-cn}
Susan Nolen-Hoeksema. 1991.
\newblock Responses to depression and their effects on the duration of depressive episodes.
\newblock \emph{J. Abnorm. Psychol.}, 100(4):569--582.

\bibitem[{Salvini et~al.(2015)Salvini, da~Silva~Dias, Lafer, and Dutra}]{DBLP:conf/medinfo/SalviniDLD15}
Rogerio Salvini, Rodrigo da~Silva~Dias, Beny Lafer, and In{\^{e}}s Dutra. 2015.
\newblock \href {https://doi.org/10.3233/978-1-61499-564-7-741} {A multi-relational model for depression relapse in patients with bipolar disorder}.
\newblock In \emph{{MEDINFO} 2015: eHealth-enabled Health - Proceedings of the 15th World Congress on Health and Biomedical Informatics, S{\~{a}}o Paulo, Brazil, 19-23 August 2015}, volume 216 of \emph{Studies in Health Technology and Informatics}, pages 741--745. {IOS} Press.

\bibitem[{Song et~al.(2020)Song, Tan, Qin, Lu, and Liu}]{DBLP:conf/nips/Song0QLL20}
Kaitao Song, Xu~Tan, Tao Qin, Jianfeng Lu, and Tie{-}Yan Liu. 2020.
\newblock \href {https://proceedings.neurips.cc/paper/2020/hash/c3a690be93aa602ee2dc0ccab5b7b67e-Abstract.html} {Mpnet: Masked and permuted pre-training for language understanding}.
\newblock In \emph{Advances in Neural Information Processing Systems 33: Annual Conference on Neural Information Processing Systems 2020, NeurIPS 2020, December 6-12, 2020, virtual}.

\bibitem[{Team(2025)}]{gemmateam2025gemma3technicalreport}
Gemma Team. 2025.
\newblock \href {https://arxiv.org/abs/2503.19786} {Gemma 3 technical report}.
\newblock \emph{Preprint}, arXiv:2503.19786.

\bibitem[{Tsakalidis et~al.(2022)Tsakalidis, Nanni, Hills, Chim, Song, and Liakata}]{tsakalidis-etal-2022-identifying}
Adam Tsakalidis, Federico Nanni, Anthony Hills, Jenny Chim, Jiayu Song, and Maria Liakata. 2022.
\newblock \href {https://doi.org/10.18653/v1/2022.acl-long.318} {Identifying moments of change from longitudinal user text}.
\newblock In \emph{Proceedings of the 60th Annual Meeting of the Association for Computational Linguistics (Volume 1: Long Papers)}, pages 4647--4660, Dublin, Ireland. Association for Computational Linguistics.

\bibitem[{Vaswani et~al.(2017)Vaswani, Shazeer, Parmar, Uszkoreit, Jones, Gomez, Kaiser, and Polosukhin}]{10.5555/3295222.3295349}
Ashish Vaswani, Noam Shazeer, Niki Parmar, Jakob Uszkoreit, Llion Jones, Aidan~N. Gomez, \L{}ukasz Kaiser, and Illia Polosukhin. 2017.
\newblock Attention is all you need.
\newblock In \emph{Proceedings of the 31st International Conference on Neural Information Processing Systems}, NIPS'17, page 6000–6010, Red Hook, NY, USA. Curran Associates Inc.

\bibitem[{Visted et~al.(2018)Visted, V{\o}llestad, Nielsen, and Schanche}]{Visted2018-si}
Endre Visted, Jon V{\o}llestad, Morten~Birkeland Nielsen, and Elisabeth Schanche. 2018.
\newblock Emotion regulation in current and remitted depression: A systematic review and meta-analysis.
\newblock \emph{Front. Psychol.}, 9.

\bibitem[{Yang et~al.(2024)Yang, Yang, Hui, Zheng, Yu, Zhou, Li, Li, Liu, Huang, Dong, Wei, and Lin}]{yang2024qwen2technicalreport}
An~Yang, Baosong Yang, Binyuan Hui, Bo~Zheng, Bowen Yu, Chang Zhou, Chengpeng Li, Chengyuan Li, Dayiheng Liu, Fei Huang, Guanting Dong, Haoran Wei, and Huan Lin. 2024.
\newblock \href {https://arxiv.org/abs/2407.10671} {Qwen2 technical report}.
\newblock \emph{Preprint}, arXiv:2407.10671.

\bibitem[{Yin et~al.(2022)Yin, Yu, Wu, Jiang, Liu, Ren, and Dai}]{DBLP:conf/ccscw/YinYWJLRD22}
Wenyi Yin, Chenghao Yu, Pianran Wu, Wenxuan Jiang, Youzhe Liu, Tianqi Ren, and Weihui Dai. 2022.
\newblock \href {https://doi.org/10.1007/978-981-99-2356-4\_15} {An intelligent mobile system for monitoring relapse of depression}.
\newblock In \emph{Computer Supported Cooperative Work and Social Computing - 17th {CCF} Conference, ChineseCSCW 2022, Taiyuan, China, November 25-27, 2022, Revised Selected Papers, Part {I}}, volume 1681 of \emph{Communications in Computer and Information Science}, pages 182--193. Springer.

\end{thebibliography}

\appendix

\section{A Note on Terminologies}
\label{app: Terminologies}

Depression relapse literature seems to have no consensus regarding the exact definitions of the terms related to the phases a person undergoes during the treatment of depression as depicted in \autoref{fig:relapse}. \citet{frank_conceptualization_1991} attempted to standardize the definitions but left it open for empirical validations. Later \citet{De_Zwart2019-pp} in their systemic review of relevant literature found that the only consensus reached till now is regarding the minimum duration required for defining an episode. 

Thus in this work we take the liberty to use the term \textit{relapse} as an umbrella term to refer to both \textit{relapse} as well as \textit{recurrence}. For similar reasons we use the term \textit{remission} to refer to both \textit{remission} as well as \textit{recovery}. More concretely, we consider a user is in remission when we come across any post in which she mentions feeling better and/or asymptomatic. Conversely, we consider a user in relapse when she explicitly mentions about the return of symptoms and feeling worse than before. Moreover, this simplifying assumption also makes it easier to work with social media posts.

\section{Pattern List}
\label{app:patterns}
\subsection{Depression}
Following is a partial list; the full list can be accessed here\footnote{\url{https://github.com/saprativa/ReDepress}}.

\texttt{``depression'',
``major depression'',
``depressive disorder'',
``dysthymia'',
``premenstrual dysphoric disorder'',
``chronic depression'',
``clinical depression'',
``depressions'',
``depressive'',
``depressive illness'',
``depressive neuroses'',
``depressive neurosis'',
``depresssion'',
``disorder depressive'',
``disorder dysphoric premenstrual'',
``disorder premenstrual dysphoric'',
``disthymia'',
``distimea'',
``distimia'',
``double depression'',
``dpression'',
``dysthymia'',
``dysthymia disorder'',
``dysthymic'',
``dysthymic dis'',
``dysthymic disorder'',
``dystimea'',
``dystimia'',
``late luteal phase dysphoric disorder'',
``llpdd'',
``major depression'',
``major depression disorder'',
``major depressive dis'',
``mdd'',
``major depressive illness'',
``premenstrual dysphoric syndrome'',
``reactive depression'',
``recurrent depressive disorder'',
``sever depression'',
``severe depression'',
``severe depressive'',
``unipolar depression'',
``unipolar depressive disorder'',
``unipolar major depression'',
``Mild mood disturbance'',
``Borderline clinical depression'',
``Moderate depression'',
``Extreme depression'',
``Minimal depression'',
``Mild depression'',
``Moderately severe depression'',
``mildly depressed'',
``moderately depressed'',
``mildly depressive'',
``moderately depressive''
}


\subsection{Diagnosed}
\begin{flushleft}
\texttt{``diagnose'', ``diagnosed'', ``diagnosing'', ``diagnosis'', ``diagnoses'', ``prescribed'', ``prescribe'', ``prescription'', ``treated'', ``treatment'', ``treating''}
\end{flushleft}

\section{Prompts}
\label{app:prompts}

\begin{tcblisting}{breakable,
  colback=blue!5!white,    % Background color of the box
  colframe=blue!75!black, % Border color of the box
  sharp corners,          % Square corners (optional)
  listing only,
  title=Llama 3.2 3B    % Title of the box
}
system_prompt = """
### Task: Please help me find the following information only about the author from its post.
[output format]:
Question 1. What is the main topic of the post?
    Answer: [Main topic identified from the post content.]
    Reason: [Reasoning based on how the main topic is inferred from the post content.]
Question 2. Does the author mention any mental health conditions if they are dealing with (only author's mental health conditions)?
    Answer: mental health condition1, mental health condition2, or Not mentioned
    Reason: [Reasoning based on how these mental health conditions are inferred from the post content.]
Question 3. Give mental health symptoms the author has experienced?
    Answer: [List of symptoms] or Not mentioned.
    Reason: [Reasoning based on how these symptoms are inferred from the post content.]
Question 4. Has the author shared any life experiences or events that have positively impacted their mental health?
    Answer: [List of experiences/events] or Not mentioned
    Reason: [Reasoning based on how these events are linked to the author’s mental health.]
Question 5. Has the author shared any life experiences or events that have negatively impacted their mental health?
    Answer: [List of experiences/events] or Not mentioned
    Reason: [Reasoning based on how these events are linked to the author’s mental health.]
Question 6. Does the author state whether they have been diagnosed with these conditions?
    Answer: mental health condition1[Yes/No or Not mentioned], mental health condition2[Yes/No or Not mentioned]
    Reason: [Reasoning based on how the diagnosis is inferred from the post content.]
Question 7. Was the diagnosis of author for mental health conditions very recent from the post creation date?
    Answer: mental health condition1[Yes/No or Not mentioned], mental health condition2[Yes/No or Not mentioned]
    Reason: [Reasoning based on how the recency of the diagnosis is inferred from the post content.]
Question 8. Has the author mentioned any therapy, counseling, or support programs they are undergoing for mental health conditions?
    Answer: Yes[Type of therapy/support program] or Not mentioned
    Reason: [Reasoning based on how therapy or counseling is inferred from the post content.]
Question 9. Has the author mentioned any medication they are taking for mental health conditions?
    Answer: Yes[Name or type of medication] or Not mentioned
    Reason: [Reasoning based on how medication use is inferred from the post content.]
Question 10. Has the author mentioned recovery of any mental health condition?
    Answer: mental health condition1[Yes/No or Not mentioned], mental health condition2[Yes/No or Not mentioned]
    Reason: [Reasoning based on how recovery is inferred from the post content.]
Question 11. Has the author mentioned supportive factors or positive influences for the recovery in their mental health condition?
    Answer: [List of supportive factors or positive influences] or Not mentioned
    Reason: [Reasoning based on how the Supportive factors or positive influences for recovery are inferred from the post content.]
Question 12. Has the author experienced a relapse of any mental health condition?
    Answer: mental health condition1[Yes/No or Not mentioned], mental health condition2[Yes/No or Not mentioned]
    Reason: [Reasoning based on how the relapse is inferred from the post content.]
Question 13. Has the author mentioned triggers or reasons for the relapse?
    Answer: [List of triggers] or Not mentioned
    Reason: [Reasoning based on how the triggers or reasons for relapse are inferred from the post content.]
[End of output format]
Please provide the answers and reasoning of these 13 questions only and exactly as in the [output format]."""
\end{tcblisting}

\begin{tcblisting}{breakable,
  colback=blue!5!white,    % Background color of the box
  colframe=blue!75!black, % Border color of the box
  sharp corners,          % Square corners (optional)
  listing only,
  title=Llama 3.1 70B    % Title of the box
}
system_prompt = """
### Task: Please help me find the following information only about the author from its post.
[output format]:
Question 1. What is the main topic of the post?
    Answer: [Main topic identified from the post content.]
    Reason: [Reasoning based on how the main topic is inferred from the post content.]
    Post Part: [Give that part of post content.]
Question 2. Does the author mention any mental health conditions if they are dealing with (only author's mental health conditions)?
    Answer: mental health condition1, mental health condition2, or Not mentioned
    Reason: [Reasoning based on how these mental health conditions are inferred from the post content.]
    Post Part: [Give that part of post content.]
Question 3. Give mental health symptoms the author is currently experiencing?
    Answer: [List of symptoms] or Not mentioned.
    Reason: [Reasoning based on how these symptoms are inferred from the post content.]
    Post Part: [Give that part of post content.]
Question 4. Has the author shared any life experiences or events that have positively impacted their mental health?
    Answer: [List of positive experiences or events] or Not mentioned
    Reason: [Reasoning based on how these positive experiences linked to the author’s mental health.]
    Post Part: [Give that part of post content.]
Question 5. Has the author shared any life experiences or events that have negatively impacted their mental health?
    Answer: [List of negative experiences or events] or Not mentioned
    Reason: [Reasoning based on how these negative experiences linked to the author’s mental health.]
    Post Part: [Give that part of post content.]
Question 6. Does the author state whether they have been diagnosed with these conditions?
    Answer: mental health condition1[Yes/No or Not mentioned], mental health condition2[Yes/No or Not mentioned]
    Reason: [Reasoning based on how the diagnosis is inferred from the post content.]
    Post Part: [Give that part of post content.]
Question 7. Was the diagnosis of author for mental health conditions very recent from the post creation date?
    Answer: mental health condition1[Yes/No or Not mentioned], mental health condition2[Yes/No or Not mentioned]
    Reason: [Reasoning based on how the recency of the diagnosis is inferred from the post content.]
    Post Part: [Give that part of post content.]
Question 8. Has the author mentioned any therapy, counseling, or support programs they are currently undergoing for mental health condition?
    Answer: Yes[Type of therapy/support program] or Not mentioned
    Reason: [Reasoning based on how therapy, counseling, or support programs is inferred from the post content.]
    Post Part: [Give that part of post content.]
Question 9. Has the author mentioned any medication they are currently taking for mental health conditions?
    Answer: Yes[Name or type of medication] or Not mentioned
    Reason: [Reasoning based on how medication use is inferred from the post content.]
    Post Part: [Give that part of post content.]
Question 10: Has the author mentioned any current changes to their medication regimen (e.g., stopping medication, starting a new medication, changing medication, increasing dosage, decreasing dosage)?
    Answer: Yes [description of change] or Not mentioned
    Reason: [Reasoning based on how these changes are inferred from the post content.]
    Post Part: [Give that part of post content.]
Question 11. Is the author currently experiencing (not in past) recovery or remission of any mental health condition?
    Answer: mental health condition1[Yes/No or Not mentioned], mental health condition2[Yes/No or Not mentioned]
    Reason: [Reasoning based on how recovery or remission is inferred from the post content.]
    Post Part: [Give that part of post content.]
Question 12. Has the author mentioned any current supportive factors or positive influences aiding their recovery or remission of their mental health condition?
    Answer: [List of supportive factors or positive influences] or Not mentioned
    Reason: [Reasoning based on how the Supportive factors or positive influences are inferred from the post content.]
    Post Part: [Give that part of post content.]
Question 13. Is the author currently experiencing (not in past) relapse or recurrence of any mental health condition?
    Answer: mental health condition1[Yes/No or Not mentioned], mental health condition2[Yes/No or Not mentioned]
    Reason: [Reasoning based on how the current relapse or recurrence is inferred from the post content.]
    Post Part: [Give that part of post content.]
Question 14. Has the author mentioned any triggers or reasons for the current relapse or recurrence?
    Answer: [List of triggers or reasons] or Not mentioned
    Reason: [Reasoning based on how the triggers or reasons are inferred from the post content.]
    Post Part: [Give that part of post content.]
[End of output format]
Please provide the answers and reasoning of these 14 questions only and exactly as in the [output format]. Answer only what is asked and do not provide any extra questions, information, explanations."""

\end{tcblisting}

\section{Annotation Guidelines}
\label{app:Annotation Guidelines}

\textit{Reproduced below is an abridged version of the annotation guidelines for the sake of brevity.} 

Each one of you will be provided with a set of excel files. Each file corresponds to one Reddit user and has the following columns:

\begin{enumerate}
    \item post\_id: A unique alphanumeric identifier for the post.
    \item author: Name of the Reddit user (mapped to a unique number for privacy preservation).
    \item created\_readable: The date and time when the post was created.
    \item title: The title of the post.
    \item selftext: The body of the post.
    \item Remission/Relapse: A drop-down with the following options:
        \begin{enumerate}
            \item Remission
            \item Recovery
            \item Relapse
            \item Other
        \end{enumerate}
    \item Memory bias: A drop-down with the following options:
        \begin{enumerate}
            \item Positive
            \item Negative
            \item No bias
        \end{enumerate}
    \item Attention bias: A drop-down with the following options:
        \begin{enumerate}
            \item Positive
            \item Negative
            \item No bias
        \end{enumerate}
    \item Interpretation bias: A drop-down with the following options:
        \begin{enumerate}
            \item Positive
            \item Negative
            \item No bias
        \end{enumerate}
    \item Rumination: A drop-down with the following options:
        \begin{enumerate}
            \item Brooding
            \item Reflection
            \item No rumination
        \end{enumerate}
\end{enumerate}

\begin{itemize}
    \item Each row in a file corresponds to a separate post. The posts are arranged in chronological order starting from old to recent.
    
    \item Your task is to look at each post title and post body and then annotate the columns Remission/Relapse, Memory bias, Attention bias, Interpretation bias and Rumination.
    
    
    
    \item Take the Dominant Label Approach: Read the entire post and decide which type of Label is most prominent overall. Select only one label.
\end{itemize}

\subsection*{Summary of Terms}

\begin{itemize}
  \item \textbf{Rumination (Thinking Over and Over)}
  \begin{itemize}
    \item \textbf{No Rumination:} The person doesn’t keep thinking about events.
    \item \textbf{Brooding:} The person is stuck in negative thoughts (e.g., ``Why do bad things always happen to me?'').
    \item \textbf{Reflection:} The person analyzes past events to understand and learn (e.g., ``What can I do differently next time?'').
  \end{itemize}

  \item \textbf{Memory Bias (What We Remember)}
  \begin{itemize}
    \item \textbf{Positive:} The person remembers things in a happy or hopeful way.
    \item \textbf{Negative:} The person remembers things in a sad or pessimistic way.
    \item \textbf{No bias:} The person recalls things in a neutral way.
  \end{itemize}

  \item \textbf{Attention Bias (What We Notice More at Present)}
  \begin{itemize}
    \item \textbf{Positive:} The person focuses more on the good in a situation.
    \item \textbf{Negative:} The person focuses more on the bad.
    \item \textbf{No bias:} The person gives equal attention to good and bad things.
  \end{itemize}

  \item \textbf{Interpretation Bias (How We Interpret Events)}
  \begin{itemize}
    \item \textbf{Positive:} Interprets ambiguous situations in an optimistic way.
    \item \textbf{Negative:} Interprets ambiguous situations as negative or unfavorable.
    \item \textbf{No bias:} Interprets ambiguous situations realistically, without excessive optimism or pessimism.
  \end{itemize}

\end{itemize}

\section{Examples of Remission and Relapse Posts}
\label{app:examples}

A typical remission post looks like the following (paraphrased for privacy reasons):

\begin{quote}
    \textit{I've been dealing with depression for a decade, but over the past few months, things have started to improve. This year has still been rough and delayed a lot of my goals by at least half a year, but my emotional responses to setbacks have become more manageable. For instance, I recently got a fine I can’t afford, but instead of spiraling, I just thought, "this is frustrating, but I’ll handle it." Even though my boss can be overbearing, I just let it go and remind myself I only interact with her once a week.}
    
    \textit{It's strange not to feel overwhelmed by everyday stressors. Is this how mentally healthy people normally cope? It's kind of unsettling to respond to difficulties without the usual emotional chaos. That internal voice of depression and anxiety still tries to sound the alarm, like “you should be panicking right now,” but it’s becoming easier to quiet that voice—even if it still feels oddly unfamiliar.}
\end{quote}
While, a relapse post looks like this:

\begin{quote}
    \textit{A bit of context: I began therapy and medication about three years ago to address depression, PTSD, and social anxiety. It took some trial and error, but eventually we found a combination that worked—Seroquel XR (twice a day), Viibryd in the morning, and Klonopin as needed.}

    \textit{Things improved significantly over the next two years, and we gradually reduced my medications to just a nightly dose of Seroquel XR. Around that time, we shifted focus to managing my ADHD. I tried Concerta without much success, and some doses of Adderall XR that worked mildly. I finally voiced that something still didn’t feel quite right.}

    \textit{We then switched from Adderall XR to IR for better control. For the past three months, I’ve been taking 15mg three times a day. It’s been helping a lot with focus—sometimes I even skip the third dose.}

    \textit{But about two weeks ago, I began feeling a low mood creeping back in. Now it’s hit hard—I’ve been emotionally withdrawn from my family, eating irregularly, and isolating myself after work. I mostly distract myself with games, movies, or music while tuning everything else out.}

    \textit{I’m wondering if the Adderall could be contributing to this downturn. It’s been effective for my ADHD, and I’d hate to start the search for a new med again. I’m not entirely opposed to restarting an SSRI, but I remember feeling emotionally flat when I was on Viibryd, even though it kept the depression in check.}
\end{quote}

\section{Distribution of Relapse Cases}
\label{app:distribution}
Relapse rates of 50\% after first, 80\% after second and 90\% after third episodes of depression are based on clinical studies. These clinical studies focus specifically on relapse rates in clinical settings, often involving hospitalized patients or those actively engaged in clinical treatments. Our dataset, derived from social media posts, reflects self-disclosed mental health statuses from a non-clinical, online environment. Consequently, our reported relapse rate (approximately 40\% in the annotated dataset: 83 relapse vs. 121 non-relapse cases) emerges organically from users' self-disclosures and subsequent clinical annotations. This rate, while similar to clinical findings, is also expected to deviate from clinical studies due to the fact that social media users are not obligated to self-report every recovery and relapse. And this is true for all social media based mental health datasets including ours. We discuss in detail the implications of using social media data for depression relapse detection in the limitations section of our paper.

Notwithstanding the above, we took extreme care during the manual filtering stages of our data curation pipeline to include only those cases which were corresponding to the first relapse after remission from depression (as much as was possible with social media self disclosures) to get unbiased data.

\section{Normality Test}
\label{app:normality}

To determine the appropriate statistical tests for \autoref{tab:relapse_stats}, we conducted normality tests on the variables. As shown in \autoref{tab:normality}, the Shapiro-Wilk test p-values for all features were less than 0.05, indicating significant deviations from normality. Given this non-normal distribution, we used the non-parametric Mann-Whitney U test as the primary method to assess group differences.

\begin{table}[t]
\centering
\resizebox{1\columnwidth}{!}{%
\begin{tabular}{@{}lcc@{}}
\toprule
\textbf{Feature} & \textbf{Shapiro-Wilk stat} & \textbf{Shapiro-Wilk p-value} \\ \midrule
Attention Bias & 0.967189 & 0.000109 \\
Interpretation Bias & 0.963059 & 0.000036 \\
Memory Bias & 0.963968 & 0.000046 \\
Rumination & 0.983268 & 0.015915 \\ \bottomrule
\end{tabular}%
}
\caption{Shapiro-Wilk normality test results for cognitive features. All features significantly deviate from normality, except Rumination which shows marginal deviation.}
\label{tab:normality}
\end{table}

\section{Average}
\label{app:average}

\begin{table}[ht]
\centering
\resizebox{0.8\columnwidth}{!}{%
\begin{tabular}{@{}lcc@{}}
\toprule
\textbf{Feature} & \textbf{Label} & \textbf{Mapped Value} \\ \midrule
\multirow{3}{*}{Memory Bias}       & Positive  & 1  \\
                                   & Negative  & -1 \\
                                   & No bias   & 0  \\ \midrule
\multirow{3}{*}{Attention Bias}    & Positive  & 1  \\
                                   & Negative  & -1 \\
                                   & No bias   & 0  \\ \midrule
\multirow{3}{*}{Interpretation Bias}    & Positive  & 1  \\
                                   & Negative  & -1 \\
                                   & No bias   & 0  \\ \midrule
\multirow{3}{*}{Rumination}        & Reflection & 1  \\
                                   & Brooding   & -1 \\
                                   & No Rumination & 0  \\ 
\bottomrule
\end{tabular}%
}
\caption{Mapping values for different cognitive parameters.}
\label{tab:bias_mapping}
\end{table}

The averaging process ensures a balanced representation of cognitive parameter scores by incorporating multiple human annotations. For each parameter, individual annotators provided qualitative labels, which were mapped to numerical values based on predefined mappings (see \autoref{tab:bias_mapping}). The final score for each of dimensions was computed as the average of available human annotations. In the majority setting, only the majority agreeing annotations are considered for averaging.

\section{Hyperparameters}
\label{app:hyper}
 For traditional machine learning models, we performed grid search over Random Forest, Gradient Boosting, XGBoost, Logistic Regression, SVM, KNN, and Neural Network classifiers trained on aggregated cognitive features. The optimal settings were chosen using 5-fold cross-validation based on F1-score. \autoref{tab:best_hyperparameters_with_aggregations_balancing} reports the best hyperparameters along with the corresponding feature aggregation strategies and class balancing methods (random oversampling or undersampling).

For transformer-based temporal models, we implemented a transformer encoder over post-level embeddings (MentalBERT, MentalRoBERTa, MPNet) enriched with cognitive markers. A wide grid of parameters was explored, including hidden dimension size, number of heads, number of layers, dropout, learning rate, batch size, weight decay, and early stopping patience. The best settings for each embedding configuration (text only, text + markers, joint embeddings) were identified via early stopping on the validation set and are summarized in \autoref{tab:best_hyperparameters_transformers}.

\begin{table*}[!ht]
\centering
\resizebox{\textwidth}{!}{%
\begin{tabular}{@{}lclll@{}}
\toprule
\textbf{Model} & \textbf{Aggregations} & \textbf{Num Features} & \textbf{Balancing} & \textbf{Best Hyperparameters} \\  
\hline
RandomForest & mean & 4 & random\_oversample & \makecell[l]{clf\_\_bootstrap=True, clf\_\_max\_depth=2, \\ clf\_\_min\_samples\_leaf=4, clf\_\_min\_samples\_split=2, \\ clf\_\_n\_estimators=100} \\  
\hline
GradientBoosting & mean+min+max+median & 16 & smote & \makecell[l]{clf\_\_learning\_rate=0.1, clf\_\_max\_depth=3, \\ clf\_\_min\_samples\_leaf=2, clf\_\_min\_samples\_split=2, \\ clf\_\_n\_estimators=50, clf\_\_subsample=0.4} \\  
\hline
XGBoost & mean+min & 8 & smote & \makecell[l]{clf\_\_colsample\_bytree=0.8, clf\_\_learning\_rate=0.01, \\ clf\_\_max\_depth=3, clf\_\_n\_estimators=50, \\ clf\_\_subsample=0.8} \\  
\hline
LogisticRegression & mean+min & 8 & none & \makecell[l]{clf\_\_C=1, clf\_\_penalty=l1, \\ clf\_\_class\_weight=balanced, \\ } \\  
\hline
SVM & mean+max & 8 & smote & \makecell[l]{clf\_\_C=0.001, clf\_\_kernel=rbf, clf\_\_degree=2, \\ clf\_\_gamma=0.01, clf\_\_shrinking=True, \\ clf\_\_class\_weight=None} \\  
\hline
KNN & mean+min+median & 12 & random\_oversample & \makecell[l]{clf\_\_n\_neighbors=5, clf\_\_algorithm=auto, clf\_\_leaf\_size=10, \\ clf\_\_metric=euclidean, clf\_\_p=1, \\ clf\_\_weights=uniform} \\  
\hline
NeuralNetwork & min+max+median & 12 & random\_oversample & \makecell[l]{clf\_\_hidden\_layer\_sizes=(128,64,32), clf\_\_activation=tanh, \\ clf\_\_alpha=1e-07, clf\_\_learning\_rate\_init=1e-05, \\ clf\_\_momentum=0.9, clf\_\_solver=adam} \\  
\hline
\end{tabular}%
}
\caption{Best hyperparameter settings, feature aggregation strategies, and balancing methods for different models.}
\label{tab:best_hyperparameters_with_aggregations_balancing}
\end{table*}

\begin{table*}[ht]
\centering
\resizebox{1\textwidth}{!}{%
\begin{tabular}{@{}lcll@{}}
\toprule
\textbf{Embedding Model} & \textbf{Configuration} & \textbf{Balancing} & \textbf{Best Hyperparameters} \\  
\hline
mpnet & No CM & No Sampling & \makecell[l]{d\_model=256, nhead=8, num\_layers=2, dropout=0.0 \\ learning\_rate=0.001, batch\_size=16, weight\_decay=1e-4, epochs=100, early\_stopping\_patience=10} \\  
\hline
mentalbert & No CM & RandomUnderSampler & \makecell[l]{d\_model=512, nhead=8, num\_layers=2, dropout=0.5 \\ learning\_rate=0.0001, batch\_size=32, weight\_decay=1e-4, epochs=100, early\_stopping\_patience=10} \\  
\hline
mentalroberta & No CM & No Sampling & \makecell[l]{d\_model=128, nhead=8, num\_layers=2, dropout=0.0 \\ learning\_rate=0.0001, batch\_size=16, weight\_decay=1e-5, epochs=100, early\_stopping\_patience=20} \\  
\hline
mpnet & CM & RandomOverSampler & \makecell[l]{d\_model=256, nhead=8, num\_layers=2, dropout=0.0 \\ learning\_rate=0.001, batch\_size=32, weight\_decay=1e-4, epochs=100, early\_stopping\_patience=10} \\  
\hline
mentalbert & CM & RandomUnderSampler & \makecell[l]{d\_model=128, nhead=4, num\_layers=4, dropout=0.0 \\ learning\_rate=0.0001, batch\_size=32, weight\_decay=1e-4, epochs=100, early\_stopping\_patience=10} \\  
\hline
mentalroberta & CM & No Sampling & \makecell[l]{d\_model=512, nhead=8, num\_layers=4, dropout=0.0 \\ learning\_rate=0.0001, batch\_size=16, weight\_decay=1e-5, epochs=100, early\_stopping\_patience=20} \\  
\hline
mpnet & CM-Emb & RandomUnderSampler & \makecell[l]{d\_model=128, nhead=4, num\_layers=2, dropout=0.0 \\ learning\_rate=0.001, batch\_size=8, weight\_decay=1e-4, epochs=100, early\_stopping\_patience=15} \\  
\hline
mentalbert & CM-Emb & RandomUnderSampler & \makecell[l]{d\_model=256, nhead=4, num\_layers=2, dropout=0.0 \\ learning\_rate=0.001, batch\_size=16, weight\_decay=1e-5, epochs=100, early\_stopping\_patience=20} \\  
\hline
mentalroberta & CM-Emb & RandomUnderSampler & \makecell[l]{d\_model=512, nhead=8, num\_layers=2, dropout=0.0 \\ learning\_rate=0.0001, batch\_size=32, weight\_decay=1e-5, epochs=100, early\_stopping\_patience=15} \\  
\bottomrule
\end{tabular}%
}
\caption{Best hyperparameter settings and balancing strategies for transformer-based temporal sequence models under different input configurations. Classification was performed using the last hidden state. Configuration short forms: No CM = No Cognitive Markers, CM = Cognitive Markers, CM-Emb = Cognitive Markers Embedding.}
\label{tab:best_hyperparameters_transformers}
\end{table*}

\section{Information Imbalance}
\label{app:info_imbalance}

\autoref{tab:info_imbalance_ablation} provides the full ablation results for each type of bias and aggregation (min, max, mean, median) for the Information Imbalance metric.

\begin{table*}[ht]
\centering
\resizebox{0.7\textwidth}{!}{%
\begin{tabular}{@{}l l c c c@{}}
\toprule
\textbf{Feature} & \textbf{Statistic} & \textbf{II$_\mathrm{Ablated}$} & \boldmath{$\Delta$} \textbf{II} & \textbf{Interpretation} \\
\midrule
Rumination          & Mean   & 0.611 & +0.014  & Most important: removing increases II \\
Interpretation Bias  & Mean   & 0.605 & +0.008  & Some importance                      \\
Attention Bias       & Mean   & 0.605 & +0.007  & Some importance                      \\
Attention Bias       & Median & 0.602 & +0.004  & Some importance                      \\
Memory Bias          & Mean   & 0.598 & -0.000  & Neutral/redundant                     \\
Rumination           & Median & 0.595 & -0.003  & Neutral/redundant                     \\
Rumination           & Min    & 0.595 & -0.003  & Neutral/redundant                     \\
Interpretation Bias  & Min    & 0.593 & -0.005  & Neutral/redundant                     \\
Memory Bias          & Max    & 0.588 & -0.010  & Slightly redundant                    \\
Interpretation Bias  & Max    & 0.587 & -0.011  & Slightly redundant                    \\
Memory Bias          & Median & 0.586 & -0.011  & Slightly redundant                    \\
Memory Bias          & Min    & 0.584 & -0.014  & Slightly redundant                    \\
Attention Bias       & Min    & 0.583 & -0.015  & Slightly redundant                    \\
Rumination           & Max    & 0.578 & -0.020  & Slightly redundant                    \\
Attention Bias       & Max    & 0.578 & -0.020  & Slightly redundant                    \\
Interpretation Bias  & Median & 0.573 & -0.025  & Slightly redundant                    \\
\bottomrule
\end{tabular}%
}
\caption{Information Imbalance ablation analysis for each feature. Positive $\Delta$II indicates the feature is highly informative for relapse detection; negative or near-zero values indicate redundancy or low unique contribution.}
\label{tab:info_imbalance_ablation}
\end{table*}

\section{Cognitive Classification Prompt}
\label{app:cognitive_prompt}

\begin{tcblisting}{breakable,
  colback=blue!5!white,    % Background color of the box
  colframe=blue!75!black, % Border color of the box
  sharp corners,          % Square corners (optional)
  listing only,
  title=   % Title of the box
}
# User prompt with post content
user_prompt = f"### Post creation date: {row['created_readable']} \n### Post: {row['title']}\n {row['selftext']}\nEnd of the post"

# System prompt for cognitive dimension classification
system_prompt = """ 
### Task: Please help me classify the following Reddit post into cognitive dimensions. 
Focus only on the author’s perspective expressed in the post. 
Assign exactly one label per dimension.

### output format:
Attention Bias:
    Answer: positive / negative / no bias
    Reason: [Reasoning based on how the bias is inferred from the post content.]

Memory Bias:
    Answer: positive / negative / no bias
    Reason: [Reasoning based on how the bias is inferred from the post content.]

Interpretation Bias:
    Answer: positive / negative / no bias
    Reason: [Reasoning based on how the bias is inferred from the post content.]

Rumination:
    Answer: reflection / brooding / no rumination
    Reason: [Reasoning based on how rumination type is inferred from the post content.]

[End of output format]

Please provide the answers and reasoning of these 4 dimensions only and exactly as in the output format. 
Answer only what is asked and do not provide any extra questions, information, or explanations.
"""
\end{tcblisting}

\end{document}